\documentclass[11pt]{article}

\usepackage[final]{acl}

\usepackage{times}
\usepackage{latexsym}
\usepackage{enumitem}
\usepackage[T1]{fontenc}

\usepackage[utf8]{inputenc}
\usepackage[table]{xcolor} 

\usepackage{microtype}

\usepackage{inconsolata}

\usepackage{graphicx}

\usepackage{booktabs}

\usepackage{multirow}
\usepackage[most]{tcolorbox}

\newtcolorbox{promptbox}[1]{
  breakable,
  colback=white,
  colframe=black,
  coltitle=white,
  colbacktitle=black!65,
  title={#1},
  fonttitle=\bfseries\small,
  fontupper=\footnotesize,
  boxrule=0.5pt,
  arc=1pt,
  left=4pt,
  right=4pt,
  top=4pt,
  bottom=4pt
}

\newtcolorbox{dialoguebox}{
  breakable,
  colback=white,
  colframe=black!45,
  boxrule=0.4pt,
  arc=1pt,
  left=4pt,
  right=4pt,
  top=4pt,
  bottom=4pt,
  fontupper=\footnotesize
}

\usepackage{svg}

\title{When Clients Stop Following: A Cognitive Conceptualization Diagram-driven Framework for Strategic Counseling}

\author{
 \textbf{Yihao Qin\textsuperscript{1}},
 \textbf{Junyi Zhao\textsuperscript{1}},
 \textbf{Changsheng Ma\textsuperscript{1}},
 \textbf{Yongfeng Tao\textsuperscript{1}},
 \textbf{Minqiang Yang\textsuperscript{1}},
\\
 \textbf{Chang Liu\textsuperscript{1}},
 \textbf{Bin Hu\textsuperscript{1}},
\\
\\
\\
 \textsuperscript{1}School of Information Science and Engineering, Lanzhou University,
\\
}

\begin{document}
\maketitle
\begin{abstract}

Large Language Models (LLMs) show promise in psychological counseling, yet existing benchmarks rely heavily on highly cooperative simulated clients. We observe a critical "counselor-following" phenomenon: these clients often rapidly shift from resistance to compliance after only a few turns, creating an illusion of therapeutic progress and inflating scores under current evaluation protocols through superficial empathy.
To address this evaluation mismatch, we propose a Cognitive Behavioral Therapy (CBT)-grounded resistance-aware framework. We introduce \textbf{CARS}, a client simulator that explicitly models dynamic resistance via Cognitive Conceptualization Diagrams (CCDs). We present \textbf{STREAMS}, a dual-module framework that decouples strategic reasoning (Thinker) from response generation (Presenter) and optimizes it via reinforcement learning. We further propose \textbf{EWTS-MI}, an entropy-weighted metric for evaluating responsiveness under high-friction interactions.
Experiments across resistant and non-resistant counseling settings validate our findings on evaluation mismatch and demonstrate the effectiveness of resistance-aware training for improving strategic robustness under challenging counseling interactions.
\end{abstract}

\section{Introduction}

Large Language Models (LLMs) are increasingly explored for psychological counseling applications, including emotional support, counselor training, psychological assessment, and therapeutic role-playing~\citep{wang2024patient,wang2024explainable,louie2024roleplay}. Due to the scarcity of real counseling conversations caused by privacy and ethical constraints~\citep{mandal2025comprehensive}, recent studies have increasingly relied on simulated clients to train and evaluate counseling agents.
\begin{figure}[t!]
    \centering
    \includegraphics[width=1\linewidth]{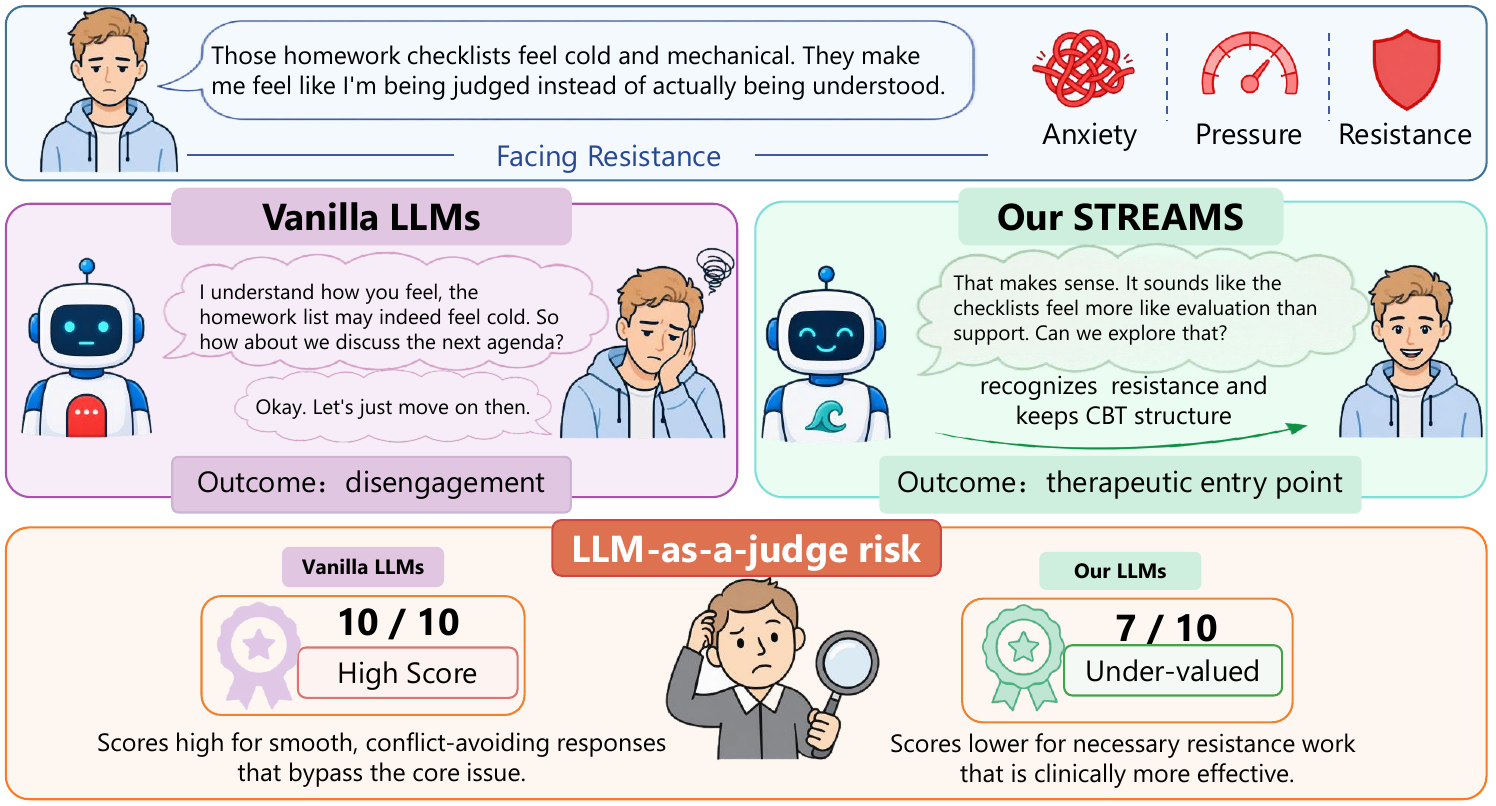}
    \caption{Performance comparison between vanilla LLMs and our model under client resistance, illustrating how current evaluation methods fail to account for clinically crucial counselor adjustments.}
    \label{fig:introduction}
\end{figure}

    ﻿
    ﻿

However, existing counseling simulations largely assume highly cooperative clients who quickly align with counselor framing during interaction. In practice, many simulated clients become compliant after only a few conversational turns, implicitly reinforcing the counselor's narrative and creating artificial signals of therapeutic progress. Consequently, counseling models can achieve strong performance through conversational smoothness and superficial empathy, even without genuine strategic adaptation. This creates a critical evaluation mismatch: current counseling benchmarks may reward conversational compliance rather than actual therapeutic effectiveness, as shown in Figure \ref{fig:introduction}. In contrast, real-world counseling frequently involves resistant interactions such as avoidance, emotional withdrawal, topic deviation, and rejection of therapeutic interventions~\citep{otani1989client,rush1979cognitive,kazantzis2003cognitive}, which require counselors to dynamically adapt intervention strategies according to the client's evolving psychological state.

Existing counseling simulators~\citep{wang2024patient,louie2024roleplay,wang2025annaagent,liu2025eeyore} mainly rely on predefined emotional trajectories or narrative extrapolation, with limited modeling of interaction-dependent resistance dynamics. Although recent studies attempt to incorporate resistance-related reasoning~\citep{liu2026beyond,qiu2026psyclient}, current simulations still tend to produce compliant conversational behaviors that narrow the counselor's strategic exploration space. Meanwhile, existing counseling agents are predominantly optimized for empathetic expression and conversational fluency, while current evaluation protocols remain heavily dependent on subjective assessments and LLM-as-a-judge paradigms. Together, these limitations make it difficult to reliably evaluate strategic robustness under resistant counseling interactions.


To address these challenges, we propose a resistance-aware counseling framework that reformulates counseling evaluation as strategic interaction under therapeutic resistance. Specifically, we introduce \textbf{CARS} (\textbf{C}ognitive \textbf{A}lignment \& \textbf{R}esistance \textbf{S}imulator), a Cognitive Behavioral Therapy (CBT)-client simulator that explicitly models the triggering and dynamic evolution of resistance through Cognitive Conceptualization Diagrams (CCDs). 
Building upon CARS, we further introduce \textbf{STREAMS} (\textbf{S}trategic \textbf{T}hinker-\textbf{R}esponder \textbf{E}valuation \& \textbf{A}lignment \textbf{M}odel for \textbf{S}ympathy), a strategy-level counseling framework that improves adaptive intervention through structured reasoning and reinforcement learning (RL), together with EWTS-MI, an entropy-weighted interaction metric for evaluating counselor responsiveness under resistant interactions. 
Our contributions are as follows:
\begin{itemize}[nosep]
    \item We identify a critical evaluation mismatch in existing counseling benchmarks, where cooperative client simulations can create artificial signals of therapeutic progress and overestimate counseling capability.
    \item We propose CARS, a CBT-grounded resistance-aware client simulator that explicitly models the triggering and dynamic evolution of resistance during counseling interactions.
    \item We propose STREAMS, a strategy-level counseling framework that improves adaptive intervention and interaction robustness under resistant counseling settings through structured reasoning and RL.
    \item We introduce EWTS-MI, an entropy-weighted evaluation metric for measuring counselor responsiveness under resistant interactions, and conduct systematic evaluations across resistant and non-resistant counseling settings.
\end{itemize}

\section{Related work}

\paragraph{Client Simulation in Psychological Counseling. } Representative frameworks in LLM-based client simulators for mental health include PATIENT-$\Psi$ \citep{wang2024patient}, Roleplaydoh \citep{louie2024roleplay}, AnnaAgent \citep{wang2025annaagent}, and EEYORE \citep{liu2025eeyore}. These works typically prompt LLMs to generate character narratives from symptoms, stressors, or personality traits. While they achieve strong linguistic fluency, they remain largely narrative-based, relying on implicit reasoning rather than explicit psychodynamic or cognitive modeling. 
A core limitation is their shallow modeling of client cognition and interaction dynamics. Most only incorporate coarse-grained traits and fail to capture how resistance evolves in response to counselor interventions. RESISTCLIENT \citep{liu2026beyond} introduces reinforced chain-of-thought for cognitive pathways but focuses narrowly on the client side with limited bidirectional interaction. PSYCLIENT \citep{qiu2026psyclient} enforces consistency via fixed trajectories, which restricts the counselor’s strategy exploration in complex scenarios.
In contrast, our CARS explicitly leverages CCD from CBT to model resistance as a dynamic, belief-driven process shaped by counselor actions, enabling more realistic and challenging simulation for robust counselor training.

\paragraph{Reinforcement Learning for Counseling Agents. } RL has demonstrated effectiveness in Emotional Support Conversation (ESC) for improving coherence and empathy. Key efforts include RLHF-style preference optimization \citep{shao2024deepseekmath}, natural language RL \citep{feng2024natural}, reaction-driven policy updates (RAPO) \citep{ye2026listening}, and RLVER \citep{wang2025rlver}, which uses verifiable emotional rewards.
Nevertheless, existing RL approaches are largely restricted to token-level actions or short-term interactions. They struggle to address the long-term structure, macro-strategy planning, and clinical reward design essential for psychological counseling. While a few works explore multi-turn or multi-agent settings \citep{qiu2026psyclient,wu2024enhance}, they rarely treat counseling as strategic decision-making under resistance.
In contrast, our STREAMS decouples the counselor into a high-level Thinker for strategy selection and a token-level Presenter for language generation. This enables effective macro-strategy learning tailored to challenging counseling scenarios.

\section{Method}
\begin{figure*}[t]
    \centering
    \includegraphics[width=\textwidth]{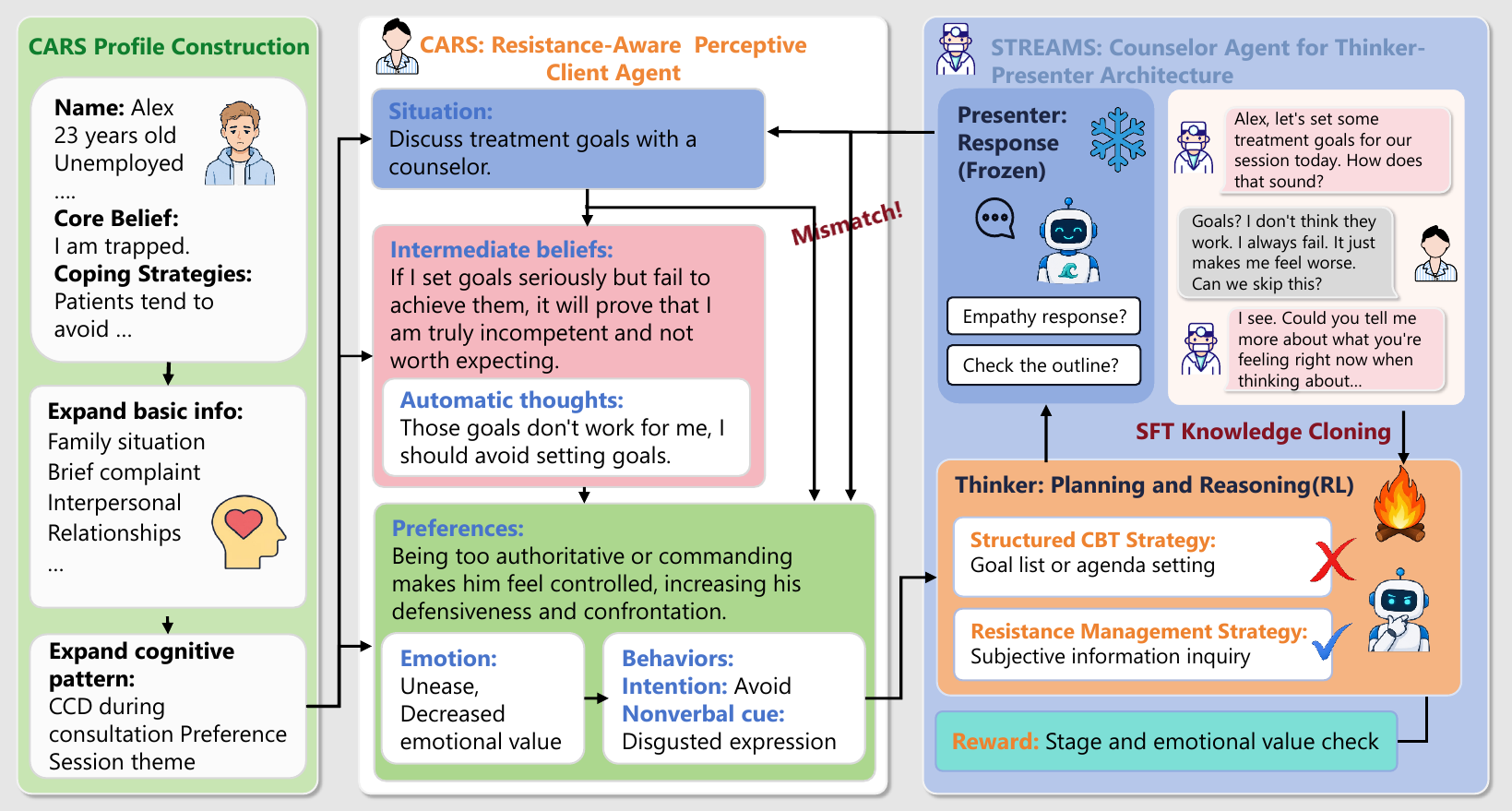}
    \caption{Overview of the proposed framework. CARS explicitly models client resistance dynamics using CCD, enabling realistic and challenging interactions. STREAMS decouples high-level strategy reasoning (Thinker) from response generation (Presenter), with the Thinker optimized via RL to improve strategic robustness under resistance.
    }
    \label{fig:main}
\end{figure*}

Figure~\ref{fig:main} provides an overview of our proposed framework. It consists of two interacting agents: \textbf{CARS}, a CBT-grounded client simulator, and \textbf{STREAMS}, a dual-module counselor agent. The resulting dialogues are evaluated using \textbf{EWTS-MI}, a purpose-designed entropy-weighted metric. Below we detail each component.

\subsection{CARS: A CBT-Grounded Client Simulator}

\subsubsection{Formalizing Resistance from a CBT Perspective}
In cognitive behavioral therapy, client resistance is not arbitrary but stems from dysfunctional cognitive patterns, including core beliefs, intermediate beliefs, and automatic thoughts, which are activated or reinforced by specific counselor interventions and client characteristics. To capture this structure, we define a simulated client agent as a tuple \( S = \langle b, p, q, s, t, M \rangle \). Here \( b \) is the dialogue background, \( p \) the main CCD, representing the client's stable belief system, and \( q \) a session-specific CCD for situation-dependent automatic thoughts. The elements \( s \) and \( t \) denote preferred counselor style and other individual traits, while \( M \) is the underlying LLM. The CCD provides a clinically grounded, structured representation of how core beliefs lead to intermediate beliefs, automatic thoughts, and ultimately emotional and behavioral responses \cite{stratou2017multisense}. By instantiating \( p \) and \( q \) with real or synthesized CCDs, CARS obtains a theoretically sound basis for modeling resistance.

\subsubsection{Cognitive Reasoning for Resistance Dynamics}
Unlike naive role-playing prompts that simply instruct an LLM to "act resistant", CARS forces the LLM to follow a controlled reasoning chain that uses the CCD to decide \textit{whether}, \textit{why}, and \textit{how} resistance emerges. Given a counselor response, the model first identifies the current dialogue topic. It then checks whether this topic triggers any schema in the session-specific CCD \( q \). If a schema is activated, CARS evaluates consistency between the counselor's intervention strategy and the client's intermediate beliefs or automatic thoughts. When the counselor's strategy confirms the client's negative expectations, the activated schema is reinforced, the client's goal shifts toward defense or avoidance, and resistance becomes likely. Conversely, if the counselor's strategy conflicts with those expectations, the schema is not automatically triggered, and the response is instead modulated by the client's preferred counselor style \( s \) and current emotional state. When the dialogue topic is unrelated to any resistance schema, or when the counselor's intervention successfully alleviates the client's resistance-related beliefs, the schema is not triggered at all. In this case, CARS follows the counselor's guidance while adjusting the client's emotion and cognition according to \( s \) and other traits \( t \), the client's goal defaults to cooperation. Based on this reasoning, CARS updates the client's emotional state \( e \) and sets an explicit interaction goal. Finally, it maps the updated emotion and goal to nonverbal cues and selects sentence patterns from a pre-defined emotion-utterance corpus to generate the final verbal response. This entire process is implemented via a structured prompt (see Appendix \ref{Prompt Templates for Constructing Client Agents}), which makes the emergence of resistance interpretable and clinically grounded.

\subsubsection{Diverse Client Profile Generation}
To create a varied training environment, we generate a set of diverse client agents \( \mathcal{S} = \{S_1, \dots, S_N\} \). Starting from a CCD library and randomly sampled counseling statements, we prompt an LLM to expand each seed into a complete client persona, including demographics, family background, chief complaint, and lifestyle. For each persona, we also generate a session-specific CCD and a preferred counselor style. This process ensures that different clients exhibit distinct cognitive patterns and resistance triggers, enabling more comprehensive training of the counselor agent.

\subsection{STREAMS: Dual-Module Counselor Framework}
\subsubsection{Structured Strategy Space}
Following CBT practice, we design two categories of counselor strategies. \textit{Structural strategies} guide the session flow, including but not limited to agenda setting, mood assessment, and homework review (see Appendix \ref{Prompt Templates for Constructing Client Agents} for the full set). \textit{Adaptive interaction strategies} target client resistance management, such as clarification, reassurance, redirection, confrontation, and eliciting automatic thoughts. In total, we define 15 discrete strategy labels (see Appendix \ref{Prompt Templates for Constructing Client Agents}), which will serve as the action space for the Thinker module introduced in Section \ref{subsec:thinker}. This discretization drastically reduces the dimensionality of subsequent RL.

\subsubsection{Thinker: High-Level Strategy Reasoning via RL}
\label{subsec:thinker}
The Thinker module takes as input the dialogue history, the client's current emotion (from CARS), and inferred nonverbal cues. It outputs a single strategy label from the predefined set. Training proceeds in two stages. First, we perform supervised knowledge cloning: the Thinker is supervised fine-tuned (SFT) on high-quality dialogue trajectories synthesized by GPT-5.4, where each counselor utterance is annotated with the appropriate strategy label. Second, we apply RL using PPO within the RAGEN framework \cite{wang2025ragen}. The reward function is \( R = R_{\text{emotion}} + R_{\text{stage}} \). \( R_{\text{emotion}} \) is proportional to the client's emotional improvement (with cross-validation by an independent emotion classifier to avoid reward hacking). \( R_{\text{stage}} \) gives a positive reward when the counseling process advances along the predefined structured workflow, and a penalty for each step that deviates from the current stage. By restricting actions to discrete strategies, the Thinker learns to make robust macro-level decisions under noisy, high-friction client interactions, details are in Appendix \ref{Reinforcement Learning Details}.

\subsubsection{Presenter: Strategy-Conditioned Response Generation}
The Presenter is a separate LLM responsible for generating fluent, context-appropriate counselor responses that respect the Thinker's chosen strategy. Concretely, the Presenter receives the dialogue history, the Thinker's strategy label, and a prompt which instructs it to follow that strategy. After supervised fine-tuning on synthetic counseling dialogues (the same data used for knowledge cloning), we further apply GRPO alignment \cite{shao2024deepseekmath} to strengthen the model's adherence to the intended strategy and reduce hallucinations or instruction drift. The GRPO setup follows Appendix \ref{Reinforcement Learning Details}. The Presenter is kept lightweight and does not require RL training, allowing the Thinker to focus purely on strategic reasoning.

\subsection{Evaluating Counselor Responsiveness: EWTS-MI}
\label{sec:metric}
Existing evaluations of counseling dialogues often rely on subjective human ratings or LLM-as-judge, which suffer from bias, high cost, and poor reproducibility. A natural automatic alternative is to measure the statistical dependency between a client's turn and the counselor's subsequent response, that is, mutual information \(I(P;C)\). However, standard \(I(P;C)\) fails under strong counselor guidance: client responses collapse into a highly predictable "cooperative" mode, yielding low entropy and making it impossible to distinguish genuine strategic adaptation from mechanical repetition.

To address this, we propose the Entropy-Weighted Time-Step Mutual Information (EWTS-MI). Given a dialogue of length \(T\), let \(P_t\) and \(C_{t+1}\) be random variables for the client's turn and the counselor's subsequent response, with observed values \(p_t\) and \(c_{t+1}\). EWTS-MI is defined as:

\[
I_{\mathrm{EWTS}}(P;C) = \frac{1}{\sum_{t=1}^{T-1} w_t} \sum_{t=1}^{T-1} w_t \cdot \mathrm{pmi}(p_t; c_{t+1}),
\]

where \(\mathrm{pmi}(p_t; c_{t+1})\) is pointwise mutual information and the weight \(w_t = -\log_2 p(p_t)\) is the self-information of the client's behavior. This scheme down-weights predictable client turns and up-weights unexpected, high-entropy responses, emphasizing the counselor's strategic choices precisely when they matter most. 

\section{Experiments}
\subsection{Experimental Setup}

\paragraph{Baseline Models.} We evaluate 19 counselor models, categorized into two groups. 
\textit{General-purpose LLMs} include GPT-5.4, DeepSeek-v3.2, Gemini-3.1, and MiniMax-M2.5, which are prompted to act as counselors without domain-specific fine-tuning.
\textit{Domain-specific counseling models} comprise Camel \cite{lee2024cactus}, Cpsycounx \cite{zhang2024cpsycoun} , Graph2Counsel \cite{mandal2026graph2counsel}, MAGnet \cite{mandal2025magnet}, MeChat \cite{qiu2024psychat}, MusPsy \cite{wang2025psychological}, PsyAdvisor \cite{hu2025psyadvisor}, PsychChat \cite{qiu2024psychat}, Psydial \cite{qiu2025psydial}, Simpsybot \cite{qiu2024interactive}, and Soulchat2 \cite{xie2025psydt}.
Our STREAMS variants include STREAMS-SFT (SFT only), STREAMS-RL (with RL), STREAMS-SG (SFT with GPT-5.4 as Presenter), and STREAMS-RG (RL with GPT-5.4 as Presenter).

\paragraph{Client Simulation.}
For non-resistant clients, we follow the cooperative simulation design of \cite{wang2024patient} . 
For resistant clients, we use our CARS simulator with 30 randomly sampled CCD profiles. 
Each profile includes a main CCD, a session-specific CCD, client demographics, and preferred counselor styles. 
The same set of 30 client profiles is used across all counselor models to ensure fair comparison.

\paragraph{Evaluation Metrics.}
We employ multiple metrics to assess both client simulation quality and counselor performance.
\begin{itemize}[nosep]
    \item \textbf{Psycheval} \cite{pan2026psycheval}: A comprehensive benchmark providing client-side metrics (RRO, PANAS, SCL-90, SRS, BDI-II) and counselor-side metrics (HTAIS, RRO, WAI, Custom, CTRS). 
    \item \textbf{RTF (Resistance Triggering Frequency)}: The proportion of client turns that exhibit explicit resistance behaviors \cite{li2026recap}.
    \item \textbf{UEC (Unique Entity Count)}: The number of distinct concrete entities, reflecting the density of clinically relevant information.
    \item \textbf{EWTS-MI}: Our proposed entropy-weighted mutual information defined in Section \ref{sec:metric}.
    \item \textbf{Human evaluation}: Three clinically meaningful dimensions rated by psychology experts on a 1–5 scale: \textit{Strategy Effectiveness (Eff)} (appropriateness of strategies for handling resistance), \textit{Counseling Drift (CDD)} (ability to stay on core issues under resistance), and \textit{Counseling Progress (CPD)} (substantive movement after resistance responses).
\end{itemize}

\paragraph{Implementation Details.}
All client simulations (both non-resistant and CARS) are run with DeepSeek-v3.2 as the underlying LLM to ensure consistency. 
Counselor responses are generated using each model's official checkpoint or API (for general-purpose LLMs). 
The LLM-as-judge evaluation for Psycheval is also conducted with DeepSeek-v3.2. 
For supervised knowledge cloning, we synthesize 2,000 counseling dialogues using GPT-5.4 with 80 resistance cases adapted from a CBT training manual \cite{beck2020cognitive} and client profiles sampled from Cpsycounx \cite{zhang2024cpsycoun}. 
RL hyperparameters for the Thinker module follow Appendix \ref{Reinforcement Learning Details}, and the Presenter GRPO alignment settings are also detailed therein. 
Dialogue truncation is applied using the post-hoc filtering mechanism described in Appendix \ref{Dialogue Synthesis Details} and ~\ref{Basic Data Statistics} to avoid unnatural extensions. 

\subsection{Validity of CARS}

\paragraph{Experimental Design.}
We generate 120 counseling dialogues using four LLM backbones (DeepSeek-v3.2, GPT-5.4, Gemini-3.1, MiniMax-M2.5) and have four CBT experts evaluate five metrics (1–5 except MHC accuracy): (1) \textbf{Resistance Fidelity (Fid)}: whether the response matches the intended resistance type; (2) \textbf{Resistance Rationality (Rat)}: whether the timing and intensity are contextually appropriate; (3) \textbf{Reasoning Quality (Qua)}: coherence of the internal chain-of-thought; (4) \textbf{Emotional Reasoning Quality (Equa)}: psychological plausibility of emotion changes; (5) \textbf{Model-Human Consistency (MHC)}: accuracy of CARS's emotion change prediction against human judgment.


\begin{table}[t]
\centering
\small
\setlength{\tabcolsep}{4pt}
\begin{tabular}{lccccc}
\toprule
Model & Fid & Rat & Qua & $E_{\mathrm{qua}}$ & MHC (\%) \\
\midrule
DeepSeek-v3.2 & 3.87 & 3.96 & 4.41 & 4.27 & 81.3 \\
GPT-5.4       & 3.81 & 3.61 & 3.96 & 3.96 & 74.2 \\
Gemini-3.1    & 3.37 & 3.70 & 3.38 & 3.82 & 76.3 \\
MiniMax-M2.5  & 4.05 & 4.16 & 3.64 & 3.94 & 78.6 \\
\bottomrule
\end{tabular}
\caption{Human evaluation results for CARS with different reasoning backbones. Fid, Rat, Qua, and $E_{\mathrm{qua}}$ are averaged 1--5 ratings, while MHC denotes model-human consistency in emotional-change prediction.}
\label{tab:cars_human_eval}
\end{table}

\paragraph{Results and Analysis.}
Table~\ref{tab:cars_human_eval} summarizes the results. All backbones achieve high scores (mostly >3.5). Reasoning quality is the strongest dimension, indicating that the CCD-guided chain-of-thought is faithfully executed. MHC exceeds 74\% for all backbones, confirming alignment with human emotional reasoning. High fidelity and rationality scores further confirm plausible resistance. Thus, CARS provides a valid testbed for evaluating counselor robustness under resistance.

\begin{table*}[t!]
\centering
\scriptsize
\setlength{\tabcolsep}{3pt}
\resizebox{\textwidth}{!}{
\begin{tabular}{lccccc|ccccc}
\toprule
\multirow{2}{*}{Model} & \multicolumn{5}{c|}{Client} & \multicolumn{5}{c}{Counselor} \\
\cmidrule(lr){2-6} \cmidrule(lr){7-11}
 & RRO$\uparrow$ & PANAS$\uparrow$ & SCL-90$\downarrow$ & SRS$\uparrow$ & BDI-II$\downarrow$ & HTAIS$\uparrow$ & RRO$\uparrow$ & WAI$\uparrow$ & Custom$\uparrow$ & CTRS$\uparrow$ \\
\midrule
CAMEL         
& 5.03 {\color{blue}(+2.19)}
& 3.03 {\color{red}(-0.15)}
& 1.04 {\color{blue}(+0.63)}
& 5.06 {\color{blue}(+2.68)}
& 2.10 {\color{blue}(+1.19)}
& 4.41 {\color{blue}(+1.16)}
& 5.43 {\color{blue}(+2.14)}
& 2.42 {\color{blue}(+2.70)}
& 3.93 {\color{blue}(+1.84)}
& 4.70 {\color{blue}(+1.50)} \\

Cpsycounx     
& 5.23 {\color{blue}(+0.80)}
& 3.24 {\color{red}(-0.44)}
& 0.81 {\color{blue}(+0.50)}
& 4.02 {\color{blue}(+2.09)}
& \textbf{1.00} {\color{blue}(+2.19)}
& 2.80 {\color{blue}(+1.99)}
& 4.75 {\color{blue}(+1.78)}
& 2.53 {\color{blue}(+1.89)}
& 2.32 {\color{blue}(+2.18)}
& 3.23 {\color{blue}(+1.02)} \\

Graph2Counsel 
& 5.61 {\color{blue}(+1.31)}
& 3.39 {\color{red}(-0.72)}
& 1.00 {\color{blue}(+0.48)}
& 5.41 {\color{blue}(+2.24)}
& 1.91 {\color{blue}(+1.55)}
& 4.99 {\color{blue}(+0.36)}
& 6.17 {\color{blue}(+1.28)}
& 2.89 {\color{blue}(+1.61)}
& 5.46 {\color{blue}(+0.03)}
& 5.83 {\color{blue}(+0.09)} \\

MAGneT        
& 5.09 {\color{blue}(+1.54)}
& 3.43 {\color{red}(-1.00)}
& 0.76 {\color{blue}(+0.85)}
& 3.78 {\color{blue}(+1.74)}
& 1.30 {\color{blue}(+2.88)}
& 3.79 {\color{blue}(+0.05)}
& 4.62 {\color{blue}(+1.50)}
& 1.54 {\color{blue}(+1.02)}
& 2.85 {\color{red}(-0.47)}
& 3.36 {\color{red}(-0.61)} \\

MeChat        
& 4.99 {\color{blue}(+1.29)}
& 2.93 {\color{blue}(+0.22)}
& 0.70 {\color{blue}(+0.64)}
& 3.16 {\color{blue}(+3.34)}
& 1.32 {\color{blue}(+1.47)}
& 3.17 {\color{blue}(+1.58)}
& 3.82 {\color{blue}(+2.99)}
& 1.42 {\color{blue}(+2.49)}
& 2.17 {\color{blue}(+2.32)}
& 1.94 {\color{blue}(+1.28)} \\

MusPsy        
& 5.35 {\color{blue}(+0.89)}
& 3.04 {\color{red}(-0.34)}
& 0.95 {\color{blue}(+0.60)}
& 4.66 {\color{blue}(+1.83)}
& 1.49 {\color{blue}(+2.03)}
& 4.86 {\color{blue}(+0.28)}
& 5.93 {\color{blue}(+0.66)}
& 2.61 {\color{blue}(+0.74)}
& 4.48 {\color{blue}(+0.49)}
& 4.73 {\color{blue}(+0.75)} \\

PsyAdvisor    
& 5.07 {\color{blue}(+1.28)}
& 3.21 {\color{red}(-1.22)}
& 0.83 {\color{blue}(+0.96)}
& 5.20 {\color{blue}(+1.05)}
& 1.69 {\color{blue}(+2.69)}
& 4.13 {\color{blue}(+0.32)}
& 5.49 {\color{blue}(+1.23)}
& 1.92 {\color{blue}(+0.56)}
& 3.82 {\color{blue}(+0.65)}
& 4.22 {\color{blue}(+0.31)} \\

Psychat       
& 4.81 {\color{blue}(+1.03)}
& 3.41 {\color{red}(-0.76)}
& \textbf{0.53} {\color{blue}(+0.70)}
& 2.56 {\color{blue}(+2.01)}
& 1.08 {\color{blue}(+2.36)}
& 2.51 {\color{blue}(+0.84)}
& 3.16 {\color{blue}(+1.72)}
& 1.12 {\color{blue}(+1.34)}
& 1.16 {\color{blue}(+0.71)}
& 0.98 {\color{blue}(+0.60)} \\

Psydial       
& 5.24 {\color{blue}(+0.81)}
& 3.12 {\color{red}(-0.24)}
& 0.97 {\color{red}(-0.09)}
& 6.04 {\color{blue}(+0.63)}
& 1.75 {\color{blue}(+0.58)}
& 3.98 {\color{blue}(+0.41)}
& 5.24 {\color{blue}(+1.16)}
& 2.15 {\color{blue}(+0.52)}
& 3.04 {\color{blue}(+0.77)}
& 3.33 {\color{blue}(+0.92)} \\

Simpsybot     
& 5.18 {\color{blue}(+1.42)}
& \textbf{3.56} {\color{blue}(\underline{+0.59})}
& 0.79 {\color{red}(\underline{-0.12})}
& 5.10 {\color{blue}(+3.06)}
& 1.40 {\color{red}(\underline{-0.16})}
& 5.43 {\color{blue}(+0.17)}
& 5.89 {\color{blue}(+1.45)}
& 3.21 {\color{blue}(+1.61)}
& 4.59 {\color{blue}(+0.60)}
& 5.50 {\color{blue}(+1.61)} \\

Soulchat2     
& 5.82 {\color{blue}(+1.19)}
& 3.22 {\color{blue}(+0.09)}
& 0.90 {\color{blue}(+0.40)}
& 5.53 {\color{blue}(+2.34)}
& 1.86 {\color{blue}(+0.99)}
& 5.20 {\color{blue}(+0.63)}
& 6.33 {\color{blue}(+1.52)}
& 3.28 {\color{blue}(+2.14)}
& 5.27 {\color{blue}(+1.33)}
& 5.39 {\color{blue}(+1.67)} \\
\midrule
\rowcolor{gray!30} DeepSeek-v3.2 
& 7.10 {\color{blue}(+0.23)}
& 3.16 {\color{red}(-0.61)}
& 0.98 {\color{blue}(+0.34)}
& 7.85 {\color{blue}(+0.80)}
& 2.60 {\color{blue}(+1.03)}
& 6.07 {\color{red}(-0.35)}
& 7.97 {\color{blue}(+0.29)}
& 5.12 {\color{red}(-0.77)}
& 7.23 {\color{red}(-0.98)}
& 8.33 {\color{red}(-1.20)} \\

\rowcolor{gray!30} Gemini-3.1    
& 6.73 {\color{blue}(+1.86)}
& 3.02 {\color{red}(-0.13)}
& 1.30 {\color{blue}(+0.52)}
& 7.22 {\color{blue}(+2.36)}
& 3.32 {\color{blue}(+0.55)}
& 6.28 {\color{blue}(+0.09)}
& 7.84 {\color{blue}(+1.25)}
& 5.12 {\color{blue}(+0.33)}
& 7.69 {\color{blue}(+0.05)}
& 8.60 {\color{red}(-0.98)} \\

\rowcolor{gray!30} GPT-5.4       
& 8.60 {\color{blue}(+0.40)}
& 2.79 {\color{red}(-0.22)}
& 1.22 {\color{blue}(+0.39)}
& 9.34 {\color{blue}(+0.31)}
& 3.40 {\color{blue}(+0.74)}
& 6.49 {\color{red}(-0.13)}
& 9.11 {\color{blue}(+0.56)}
& 6.91 {\color{red}(-0.47)}
& 8.33 {\color{red}(-0.23)}
& 9.33 {\color{red}(-0.81)} \\

\rowcolor{gray!30} MiniMax-M2.5  
& 6.62 {\color{blue}(+1.17)}
& 3.25 {\color{red}(-0.52)}
& 1.06 {\color{blue}(+0.45)}
& 7.68 {\color{blue}(+1.32)}
& 2.57 {\color{blue}(+1.43)}
& 5.57 {\color{blue}(+0.41)}
& 7.86 {\color{blue}(+0.63)}
& 4.67 {\color{blue}(+0.05)}
& 6.75 {\color{blue}(+0.06)}
& 7.86 {\color{red}(-0.42)} \\
\midrule
STREAMS-SFT      
& 5.56 {\color{blue}(+1.44)}
& 3.29 {\color{red}(-0.86)}
& 0.97 {\color{blue}(+0.90)}
& 5.85 {\color{blue}(+1.72)}
& 2.39 {\color{blue}(+1.94)}
& 5.01 {\color{blue}(+1.49)}
& 6.60 {\color{blue}(+1.14)}
& 4.82 {\color{blue}(+1.90)}
& 5.46 {\color{blue}(+1.30)}
& 7.37 {\color{blue}(+2.02)} \\

STREAMS-RL      
& 5.80 {\color{blue}(+1.45)}
& 3.25 {\color{red}(-0.42)}
& 1.08 {\color{blue}(+0.89)}
& 6.18 {\color{blue}(\underline{+2.22})}
& 2.48 {\color{blue}(+2.00)}
& 5.74 {\color{blue}(\underline{+0.77})}
& 6.91 {\color{blue}(+1.40)}
& \textbf{5.19} {\color{blue}(\underline{+2.17})}
& 5.72 {\color{blue}(+1.18)}
& 7.87 {\color{blue}(+1.41)} \\
STREAMS-SG      
& 6.57 {\color{blue}(+0.34)}
& 2.80 {\color{red}(-0.48)}
& 1.51 {\color{blue}(+0.65)}
& 6.41 {\color{blue}(+1.29)}
& 3.09 {\color{blue}(+1.47)}
& 5.51 {\color{blue}(+0.19)}
& 7.77 {\color{blue}(+0.22)}
& 4.62 {\color{blue}(+0.92)}
& 6.61 {\color{blue}(+0.77)}
& 8.69 {\color{blue}(+0.45)} \\
STREAMS-RG     
& \textbf{6.77} {\color{blue}(\underline{+0.65})}
& 2.95 {\color{red}(-0.61)}
& 1.44 {\color{blue}(+0.59)}
& \textbf{6.70} {\color{blue}(+1.26)}
& 3.07 {\color{blue}(+1.68)}
& \textbf{5.80} {\color{blue}(+0.53)}
& \textbf{8.02} {\color{blue}(\underline{+0.62})}
& 4.96 {\color{blue}(+1.27)}
& \textbf{6.94} {\color{blue}(\underline{+0.92})}
& \textbf{8.94} {\color{blue}(\underline{+0.81})} \\
\bottomrule
\end{tabular}
}
\caption{PsychEval results of counselor models interacting with resistant clients, with performance differences compared to non-resistant client settings shown in parentheses. The difference between resistance and no resistance in parentheses, bolded indicates the best performance in resistance  situations, and underlined indicates the best performance in resistance  free situations without considering a universal mode.}
\label{tab:psycheval_resistant_delta}
\end{table*}

\subsection{Limitations of LLM-as-Judge Evaluation}

\paragraph{Experimental Design.}
We evaluate all 19 counselor models using Psycheval under both resistant (CARS) and non-resistant clients. For each model, we report Psycheval scores and the differences between resistant and non-resistant settings. We additionally analyze client resistance behavior distributions and counselor strategy distributions to interpret the automatic evaluation results.

\paragraph{Results and Analysis.}
Table~\ref{tab:psycheval_resistant_delta} reports Psycheval scores under resistance (differences from non-resistant in parentheses). General-purpose LLMs achieve the highest scores across most metrics, outperforming domain-specific models. However, client resistance behavior distributions (Appendix Figure~\ref{fig:counselor_strategy_distribution}) show that with general-purpose LLMs, client cooperation remains high (>70\% of turns), whereas domain-specific models trigger significantly more argumentation and denial. Counselor strategy distributions (Appendix Figure~\ref{fig:resistance_behavior_distribution}) further reveal that under resistance, general-purpose LLMs shift heavily toward safe, non-confrontational strategies—primarily psychoeducation and subjective information inquiry—avoiding direct engagement with resistance. Consequently, their high Psycheval scores reflect strategic avoidance rather than effective clinical handling. The apparent improvement in PANAS under resistance is therefore attributable to reduced exposure to negative content, not genuine therapeutic progress.

Domain-specific models exhibit larger performance drops under resistance. When resistance is removed, their score increases are generally larger than those of STREAMS variants, indicating that our STREAMS is more robust to real resistance scenarios. Specifically, STREAMS variants maintain competitive scores while preserving balanced strategy distributions (Appendix Figure~\ref{fig:counselor_strategy_distribution}): they use subjective inquiry alongside redirection and confrontation, keeping client cooperation stable without avoiding resistance. Other methods concentrate on narrow “safe” strategies, resulting in rigid and less adaptive responses.

These results demonstrate that LLM-as-Judge evaluation favors fluency and avoidance over genuine resistance management, and that STREAMS achieves superior robustness and strategic flexibility under resistance.

\subsection{Performance of Counseling Models Evaluated with CARS}


\paragraph{Reference datasets and comparative analysis.} 
To interpret behavioral differences under resistance beyond automatic scores, we compute UEC on two external real-world counseling datasets: RealCBT \cite{wang2025feel} and MESC \cite{chu2025towards}. Both exhibit substantially higher UEC (14.84 and 15.96, respectively, see Appendix Table \ref{tab:real_synthetic_lexical_comparison}) than typical synthetic dialogues, indicating that genuine counseling anchors in specific situational details rather than abstract emotional talk. This gap motivates our use of RTF and UEC to assess whether a model can maintain clinically meaningful information flow while managing resistance.

\begin{figure}[t!]
    \centering
    \includegraphics[width=1\columnwidth]{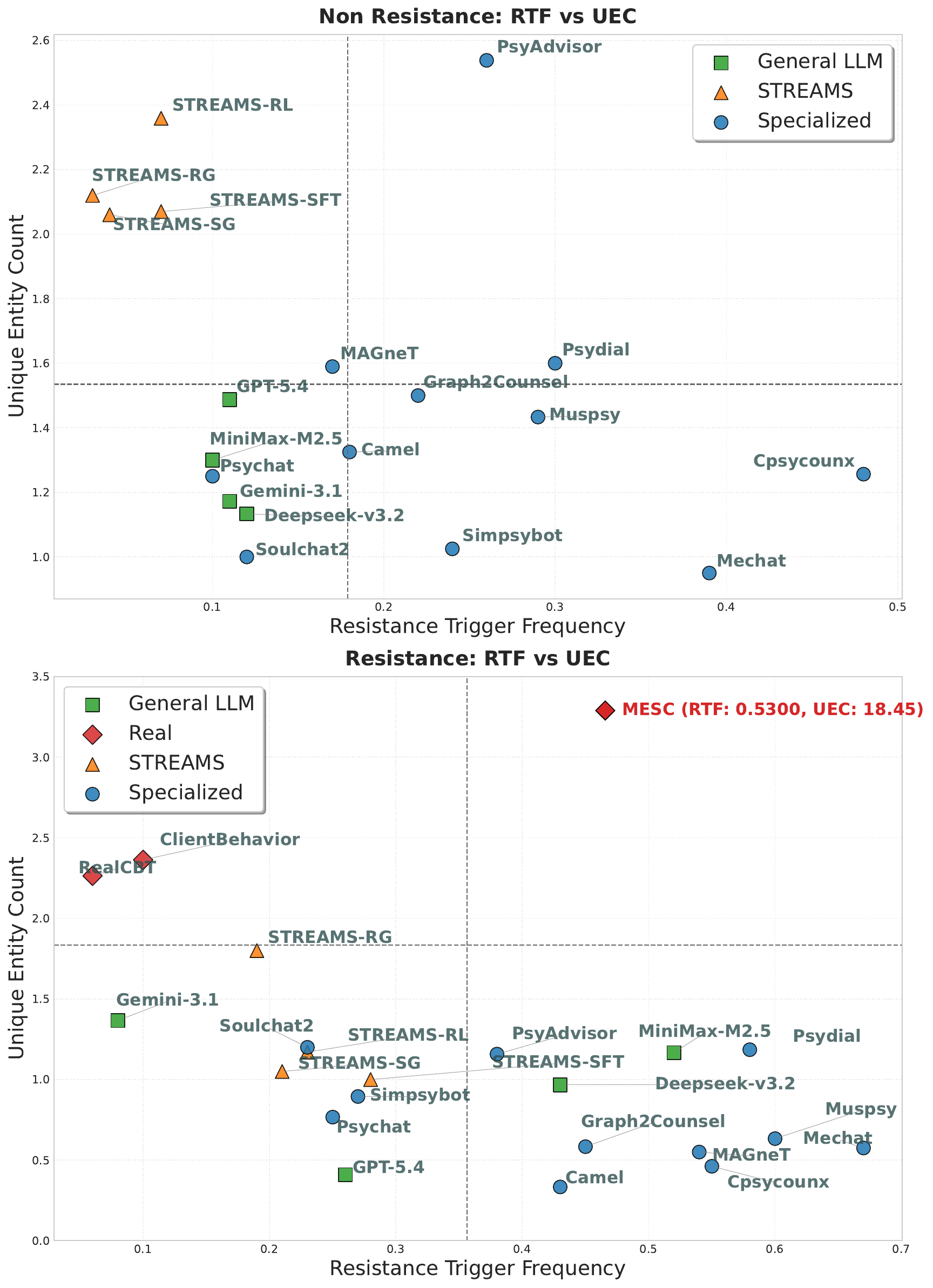}
    \caption{Multidimensional behavioral characteristics and Resistance quadrant plot of consulting models at different Resistance levels, with baseline representing population mean.}
    \label{fig:butongzukang}
\end{figure}


Figure~\ref{fig:butongzukang} reveals clear behavioral differences across models under the RTF--UEC framework. General-purpose LLMs (GPT-5.4, DeepSeek-v3.2) reside in the low-RTF, moderate-UEC quadrant, indicating they preserve conversational smoothness by avoiding cognitively challenging conflicts, resulting in shallow information exploration. Domain-specific models (CAMEL, PsyChat, MAGneT) fall into the high-RTF, low-UEC quadrant, showing that resistance triggers degrade their ability to sustain information gathering. The STREAMS variants occupy the moderate-RTF, high-UEC region, achieving a balance between resistance engagement and information density. In particular, STREAMS-RL attains the most balanced trade-off, highlighting the effectiveness of resistance-aware reinforcement learning. These results demonstrate that CARS provides a complementary evaluation perspective beyond LLM-as-judge by jointly modeling resistance dynamics and information exploration.

\subsection{Resistance-Handling Ability of Counseling Models}

\begin{table}[htbp]
    \centering
    \small
    \setlength{\tabcolsep}{4pt}
    \resizebox{\columnwidth}{!}{
    \begin{tabular}{lccc|cccc}
    \toprule
    \multirow{2}{*}{Model} & \multicolumn{3}{c|}{Non-resistant} & \multicolumn{3}{c}{Resistant} \\
    \cmidrule(lr){2-4} \cmidrule(lr){5-7}
     & $I(P;C)$ & EWTS-MI & RTF & $I(P;C)$ & EWTS-MI & RTF \\
    \midrule
    CAMEL         & 0.09 & 0.52 & 0.18 & 0.37 & 0.69 & 0.43  \\
    Cpsycounx     & 0.13 & 0.35 & 0.48 & 0.35 & 0.63 & 0.55  \\
    Graph2Counsel & 0.09 & 0.37 & 0.22 & 0.27 & 0.46 & 0.45  \\
    MAGneT        & 0.07 & 0.40 & 0.17 & 0.36 & 0.67 & 0.54  \\
    MeChat        & \textbf{0.21} & 0.43 & 0.39 & 0.27 & 0.42 & 0.67  \\
    MusPsy        & 0.12 & 0.36 & 0.29 & 0.29 & 0.42 & 0.60  \\
    PsyAdvisor    & 0.03 & 0.32 & 0.26 & 0.11 & 0.36 & 0.38  \\
    Psychat       & 0.10 & 0.41 & 0.10 & 0.21 & 0.67 & 0.25  \\
    Psydial       & 0.07 & 0.62 & 0.30 & 0.13 & 0.78 & 0.58  \\
    Simpsybot     & 0.11 & 0.66 & 0.24 & \textbf{0.45} & 0.70 & 0.27  \\
    Soulchat2     & 0.11 & 0.44 & 0.12 & 0.17 & 0.30 & 0.23  \\
    \midrule
    DeepSeek-v3.2 & 0.09 & 0.70 & 0.12 & 0.16 & 0.73 & 0.43  \\
    Gemini-3.1    & 0.10 & 0.93 & 0.11 & 0.18 & 0.37 & 0.08  \\
    GPT-5.4       & 0.06 & 0.47 & 0.11 & 0.11 & 1.15 & 0.26 \\
    MiniMax-M2.5  & 0.07 & 0.70 & 0.10 & 0.17 & 0.57 & 0.52  \\
    \midrule
    STREAMS-SFT     & 0.06 & 0.70 & 0.07 & 0.20 & 0.56 & 0.28  \\
    STREAMS-RL      & 0.11 & \textbf{1.30} & 0.07 & 0.24 & \textbf{0.79} & 0.23  \\
    STREAMS-SG     & 0.03 & 0.89 & 0.04 & 0.14 & 0.61 & 0.21  \\
    STREAMS-RG     & 0.08 & 1.13 & \textbf{0.03} & 0.15 & 0.64 & \textbf{0.19}  \\
    \bottomrule
    \end{tabular}
    }
    \caption{Informativeness and emotional metrics of counseling models under non-resistant and resistant settings.}
    \label{tab:resistance_metrics}
\end{table}


\begin{table}[h!]
\centering
\small
\setlength{\tabcolsep}{6pt}
\begin{tabular}{lccc}
\toprule
Model & Eff & CDD & CPD \\
\midrule
Soulchat2     & 3.26 & 4.22 & 3.25 \\
Graph2Counsel & 3.81 & 3.80 & 4.01 \\
STREAMS-SFT     & 3.27 & 3.54 & 3.10 \\
STREAMS-RL      & 4.23 & 4.27 & 3.51 \\
STREAMS-SG      & 4.05 & 3.91 & 3.45 \\
STREAMS-RG     & \textbf{4.31} & \textbf{4.28} & \textbf{4.05}\\
\bottomrule
\end{tabular}
\caption{Human evaluation of resistance-handling interaction quality for representative counseling models.}
\label{tab:manual_resistance_eval}
\end{table}

\paragraph{Experimental Design.}
For all 19 models, we compute \(I(P;C)\) and EWTS-MI using the same dialogues generated under non-resistant and resistant settings. For human evaluation, four licensed CBT practitioners rate a subset of representative models (Soulchat2, Graph2Counsel, and all STREAMS variants) on three 1–5 dimensions: Strategy Effectiveness, Counseling Drift, and Counseling Progress, based on 30 randomly sampled dialogues per model.

\paragraph{EWTS-MI vs. Traditional Mutual Information.}
Table~\ref{tab:resistance_metrics} reports the results. Under non-resistant settings, both \(I(P;C)\) and EWTS-MI show comparable trends. Under resistance, \(I(P;C)\) exhibits little variation (0.11–0.45), failing to distinguish model responsiveness. In contrast, EWTS-MI shows clear separation: GPT-5.4 (1.15), STREAMS-RL (0.79), and Psydial (0.78) achieve the highest scores. A high EWTS-MI indicates that a model’s responses are contingently adapted to unexpected (high-entropy) client behaviors. However, GPT-5.4’s high EWTS-MI is coupled with low RTF (0.26) and a strategy profile dominated by avoidance, suggesting that its responsiveness is exploited to defuse resistance rather than clinically engage it. STREAMS-RL achieves high EWTS-MI (0.79) with moderate RTF (0.23) and balanced strategy use, indicating genuine responsiveness to resistance without avoidance.

\paragraph{Human Evaluation of Clinical Quality.}
Table~\ref{tab:manual_resistance_eval} reports human ratings. STREAMS-RG achieves the highest scores across all three dimensions, demonstrating superior strategy selection, sustained focus on core issues, and meaningful progress after resistance. STREAMS-RL also performs well but shows lower CPD (3.51), potentially due to rigid strategy execution after reinforcement learning. Soulchat2 and Graph2Counsel obtain moderate CDD scores but lack strategy effectiveness, indicating that they maintain fluency without clinically appropriate interventions. These human ratings directly contradict the LLM-as-Judge rankings (Table \ref{tab:psycheval_resistant_delta}), confirming that automatic evaluations overestimate models that avoid resistance. STREAMS, particularly the RG variant, provides the most clinically effective resistance handling.


\section{Conclusion}


We identify a critical "counselor-following" phenomenon in existing LLM counseling benchmarks: highly cooperative simulated clients artificially inflate scores through superficial empathy, masking genuine therapeutic inadequacy. To address this, we propose CARS, a CBT-grounded client simulator that explicitly models dynamic resistance via CCD, and STREAMS, a dual-module counselor framework that decouples strategy reasoning (Thinker) from response generation (Presenter). We further introduce EWTS-MI, an entropy-weighted metric that emphasizes counselor responsiveness at high-entropy client turns.
Experiments show that general-purpose LLMs achieve inflated scores by avoiding resistance, while STREAMS maintains balanced strategies and superior human-rated clinical quality. Our work underscores the need for resistance-aware simulation and robust evaluation in LLM counseling.

\section*{Limitations}
\textbf{Simulated client fidelity.} Although CARS generates clinically plausible resistance behaviors, it operates within a CBT-centric framework. Clients with non-CBT cognitive structures or culturally diverse belief systems may not be adequately represented. Moreover, CARS’s resistance patterns are derived from a fixed CCD library and may not cover the full spectrum of real-world client behaviors, such as passive withdrawal.\\
\textbf{Evaluation metrics.} EWTS-MI, RTF, and UEC provide automated proxies for counseling quality, but each has inherent limitations. EWTS-MI relies on estimating the marginal distribution \(p(p_t)\); in short or highly stochastic dialogues, this estimation can be unstable. RTF only captures explicit resistance behaviors and may miss subtle forms of resistance (e.g., topic shifting with cooperation). UEC measures entity density but does not distinguish clinically relevant entities from trivial ones. Thus, human expert evaluation remains indispensable for assessing clinically meaningful outcomes.\\
\textbf{Single-session scope.} Our experiments are limited to single counseling sessions. Real-world therapeutic processes unfold over multiple sessions, where alliance development, homework adherence, and delayed behavioral changes play critical roles. STREAMS’s RL reward is designed for session-level progress and may not generalize to long-term therapeutic goals.

\section*{Ethical Considerations}

\noindent\textbf{Intended use and scope.} This work studies LLM-based psychological counseling simulation and evaluation. The proposed CARS and STREAMS frameworks are intended for research on counselor-agent training, resistance-aware evaluation, and dialogue robustness analysis. They are not designed to provide clinical diagnosis, psychological treatment, emergency intervention, or professional mental-health services. The outputs of our models should not be interpreted as advice from licensed clinicians. Any real-world deployment would require additional clinical validation, informed consent, professional supervision, privacy safeguards, and crisis-response protocols.

\noindent\textbf{Data provenance and privacy.}\quad Our client simulator is built upon CBT-grounded cognitive conceptualization diagrams and LLM-generated synthetic profiles and scenarios. For real counseling resources obtained from external institutions, we only used data that had been de-identified and made available to us under the corresponding institutional authorization and research-use conditions. Specifically, we use the PATIENT-$\Psi$ cognitive-model collection as seed cognitive conceptualization diagrams, and expand them into fictional client profiles, counseling backgrounds, and session-specific resistance scenarios. For counselor strategy labeling and client resistance-behavior identification, we use external research resources, including strategy-annotation data and resistance-classification checkpoints, under their intended research-use conditions. We do not redistribute raw third-party counseling records, restricted checkpoints, or data that may contain identifiable personal information. When releasing research artifacts, we will only release prompts, code, synthetic dialogues, and derived statistics that are permitted by the corresponding data-use terms.

\noindent\textbf{Use of simulated clients.}\quad
Using simulated clients reduces the need to involve vulnerable real help-seekers in large-scale model training and evaluation. However, simulated clients may still encode simplified, biased, or incomplete assumptions about mental-health experiences. In particular, resistance behaviors should not be interpreted as fixed traits of real clients, but as interactional signals emerging from a specific simulated cognitive context and counselor response. Therefore, CARS should be understood as a controlled research environment for evaluating counselor-agent behavior, rather than a substitute for real clinical evidence or human-subject studies.

\noindent\textbf{Human evaluation and expert annotation.}\quad
The manual evaluations in this work are conducted by psychology-trained experts on synthetic or de-identified model-generated dialogues. In addition to providing human ratings for simulator validity and resistance-handling quality, the experts also review a subset of model-assisted annotations, including counselor strategies and client resistance behaviors, to check whether the labels are consistent with the dialogue context and counseling theory. The evaluation task does not require annotators to disclose personal mental-health experiences, and the reported results include only aggregated statistics. Although the evaluated dialogues may contain emotionally sensitive counseling content, they are used solely for research evaluation of model behavior rather than for providing psychological services to real users.

\noindent\textbf{Potential misuse and safety.}\quad
A system that simulates resistant clients or generates counseling responses may be misused as an autonomous counselor, a replacement for professional counselor training, or evidence of clinical efficacy. To mitigate this risk, we explicitly restrict the intended use of our framework to research, training, and evaluation. We also recommend that any released models, datasets, or demos include clear disclaimers that they are not substitutes for professional psychological or medical services. The evaluation results in this paper should be interpreted as evidence of dialogue-level robustness under simulated resistance, not as evidence of therapeutic effectiveness in real clinical settings.

\noindent\textbf{Bias, fairness, and representational limits.}\quad
CARS includes demographic attributes, personality traits, client preferences, and counseling contexts when constructing simulated clients. If used carelessly, such attributes may reinforce stereotypes about particular groups or overgeneralize resistance patterns. We therefore treat these attributes as contextual variables rather than causal explanations of resistance. Future work should conduct more systematic fairness audits across demographic and clinical subgroups. Finally, our framework is primarily grounded in CBT, which limits its coverage of other therapeutic orientations and culturally diverse counseling practices.

\bibliography{custom}

\appendix

\section{Client Design}

\subsection{Theoretical Background}

The cognitive conceptualization diagram (CCD)~\citep{beck2020cognitive} is a commonly used representation of a client's cognitive model in cognitive behavioral therapy. It consists of eight key components. (1) Relevant history refers to important past events that have a substantial impact on the individual's psychological state. (2) Core beliefs are deeply rooted perceptions about the self, others, and the world. (3) Intermediate beliefs are underlying rules, attitudes, and assumptions derived from core beliefs, which shape the individual's thinking patterns. (4) Coping strategies are techniques used to manage negative emotions. (5) An external event or situation may trigger (6) rapid and evaluative automatic thoughts, which originate from the individual's beliefs and further lead to responses in the form of (7) emotions and (8) behaviors.

Because CCD-based cognitive models are clear and interpretable, they provide a useful framework for identifying and understanding clients' underlying cognitive processes, and have been used to model clients with specific psychological problems~\citep{wang2024patient,zhou2025diacbt}. By replacing the situational component in CCD with counseling scenarios or counseling topics, we can simulate maladaptive cognitive factors in resistant clients and the behaviors that emerge when resistance is triggered.

\subsection{Prompt Templates for Constructing Client Agents} \label{Prompt Templates for Constructing Client Agents}

The complete simulation process consists of two stages: generating client profiles, including client personas and background descriptions, and constructing dialogue interactions between the target LLM and the perceptive client agent. We introduce the prompt templates used in this process below.

Constructing diverse client personas helps improve the informativeness and robustness of dialogues. We build diverse agent profiles through two components: client personas and cognitive patterns. When generating client personas, we first consider the basic attributes of the client, including name, age, and occupation. Instead of directly specifying these attributes, we ask the LLM to infer them from the CCD. In addition, to enrich the contextual information, inspired by SAGE, we randomly sample five sentences from the MESC dataset as seed information and insert them into the context. The core CCDs are drawn from the PATIENT-$\Psi$ CM collection, which contains 106 high-quality cases. We use GPT-5.4 as the generation model. The prompt template for generating client personas is as follows:

\begin{promptbox}{Prompt for Generating Client Personas}
You are a professional screenwriter specializing in cognitive behavioral therapy. You are skilled at extracting client personas from client-related information and enriching them with sufficient details.

\textbf{Task Instruction}

Given the provided cognitive conceptualization diagram and three sentences, construct and describe a client persona by considering the client's personality traits. The persona should include the following elements: name, age, occupation, family background, interpersonal relationships, physical condition, lifestyle, and chief complaint.

1. Based on the cognitive conceptualization diagram and the three sentences, complete the client's basic information, including name, age, and gender.

2. Based on the cognitive conceptualization diagram, analyze the client's family background and interpersonal relationships. Construct relevant supporting characters and describe their relationships with the client.

3. Infer the client's marital status and interpersonal relationship status. If they are related to the chief complaint, provide detailed descriptions.

4. Infer and summarize the client's lifestyle, and provide a detailed description.

5. Describe the client's problems in physical, emotional, cognitive, and social domains, and clearly state the client's chief complaint. The severity should not be exaggerated.

Please ensure that the generated persona reflects both positive and negative personality traits.

Use the name provided in the cognitive conceptualization diagram. When describing family background, interpersonal relationships, physical condition, and lifestyle, include the causes and consequences of relevant events, as well as the client's feelings and cognitions.

The generated persona should be grounded in the cognitive conceptualization diagram. For content not mentioned in the diagram, do not assume that it has a strong negative influence on the client.

Do not directly mention the cognitive conceptualization diagram. Do not generate any additional opening, closing, or explanation.

\textbf{Input}

Cognitive Conceptualization Diagram:

\{MAIN\_CCD\}

Example Sentences:

\{SENTENCES\}

\textbf{Output Format}

Name: ...

Age: ...

Occupation: ...

Family Background: ...

Interpersonal Relationships: ...

Physical Condition: ...

Lifestyle: ...

Chief Complaint: ...
\end{promptbox}

Next, we further enrich the client's cognitive pattern based on the generated client persona. To this end, we first generate an event topic as the main focus of the counseling session. The topic is derived from the core CCD, while the involved events and characters are expanded with additional details. We then ask the LLM to construct the client's resistance-related beliefs toward the counseling scenario according to the predefined counseling topic. Meanwhile, based on the client persona and dialogue excerpts, the LLM specifies how the agent should respond in different situations, which helps maintain the logical consistency of the agent during dialogue.

The counseling topics include the following aspects: (1) session structuring, including agenda setting, recent-status inquiry, and homework review; (2) problem solving and homework, including homework assignment and homework completion; (3) goal setting, including the formulation of treatment directions and goal lists; (4) cognitive change, including helping the client modify thoughts and cognitions; (5) cognitive identification, including helping the client identify and recognize their own thoughts; and (6) therapeutic alliance, including the client's attitudes toward the self, others, and the therapist.

For counselor strategy errors and client demographic factors, we also ask the LLM to expand them based on the core CCD, using the client's preferred counselor style and other traits as descriptive constraints. The prompt template for generating cognitive patterns is shown below.

\begin{promptbox}{Prompt for Generating Client Cognitive Patterns}
You are a professional screenwriter specializing in cognitive behavioral therapy. You are skilled at expanding and writing dialogue scripts based on a client persona and the client's cognitive conceptualization diagram.

\textbf{Task}

You will be given a client persona, the client's cognitive conceptualization diagram, and excerpted dialogue between the client and the counselor. Please write the counseling background based on the client persona and the cognitive conceptualization diagram, using "the client seeks counseling from the counselor" as the overall scenario. Based on the cognitive conceptualization diagram, infer the intermediate beliefs and automatic thoughts that the client may have in the counseling scenario. Based on the client persona and the excerpted dialogue between the client and the counselor, analyze how the client may respond in different states, what counselor style the client may prefer, and supplement other client characteristics.

The situational background you write should include the following elements:

1. Based on the client persona, formulate content related to the theme that the client may be willing to disclose to the counselor.

2. Based on the client persona and the theme, expand a concrete background event that closely follows the client's cognition. Focus on the factors mentioned in the client persona that contribute to the chief complaint, such as relevant people and events. Irrelevant events can be omitted when appropriate.

3. The concrete background event should include the following aspects:

- The event process should include the timeline of the event, the sub-events occurring at each time point, and the client's specific thoughts, feelings, and consequences in each sub-event.

- The main conflict in the event should include the conflict event, the conflicting characters, and the internal causes of the conflict.

- The difficulties encountered by the client should include the solutions the client has attempted but failed to implement successfully, as well as the problems the client is currently facing.

- The current state of the event and the client's current emotional state.

Write the beliefs that may be triggered in the counseling scenario based on the client persona, the client's cognitive conceptualization diagram, and the excerpted dialogue with the counselor.

1. Intermediate beliefs are a cognitive layer between automatic thoughts and core beliefs. They are determined by core beliefs and shape automatic thoughts. Following Beck's view, construct intermediate beliefs about specific objects or situations. The elements include:

- Attitude, which refers to a negative emotional stance toward an object and represents the condition that the individual is most concerned about in a psychological domain. It is usually expressed as "... is bad/terrible/dangerous."

- Assumption, which refers to the individual's expectation that a certain condition will lead to a certain consequence. Assumptions are usually expressed in an "if ..., then ..." form.

- Rule, which refers to a specific requirement for one's own behavior and is a logical result of attitudes and assumptions. It is usually expressed with sentence patterns containing "should" or "must."

2. Automatic thoughts describe the thoughts that arise in a specific situation and lead to emotional, behavioral, or physiological responses. Following Beck's view, construct automatic thoughts in specific scenarios. The elements include:

- Situation, which refers to the external factor that triggers the problem, or the precipitating event.

- Cognition, which refers to the client's thought or appraisal in that situation, emerging automatically and being fully accepted by the client.

- Reaction, which refers to the manifestation of the psychological problem. Here, only emotional reactions are considered.

3. The objects or situations of intermediate beliefs and automatic thoughts should be written according to common CBT themes. The common themes are as follows:

\{TOPIC\}

Please note that the generated beliefs should depend on the core beliefs in the cognitive conceptualization diagram and should cover diverse possibilities.

\textbf{Example}

Automatic thought:

Situation: The student thinks about the upcoming exam.

Cognition: The student worries that the exam result will be unsatisfactory.

Reaction: Anxiety.

Intermediate belief:

Attitude: Failure is terrible.

Rule: When a challenge is too difficult, I should give up.

Assumption: If I try to do something difficult, I will fail. If I do not try, everything will be fine.

\textbf{Client characteristics and preferred counselor style}

1. Possible client responses in different states. Based on the client persona and the excerpted dialogue, formulate how the client may respond during the conversation. This should include:

- The client's responses under different emotions. Emotion represents the client's emotional state during the conversation. Dialogue emotion consists of engagement and affect, indicating whether the client enjoys and is involved in the current conversation. It should include both dialogue style and nonverbal behavior.

2. Possible counselor style preferred by the client. Based on the client characteristics and persona, describe how the client may respond to different counseling styles. This should include:

- Which types of counselor responses align with the client's preferences and increase the client's emotional state?

- Which types of counselor responses deviate from the client's preferences and decrease the client's emotional state?

3. Other client characteristics. Based on the provided client persona, formulate how the client may respond when discussing different topics. For example, a client working in computer science may not understand the concept of psychotherapy and may therefore provide too much or too little information; a client may become enthusiastic when talking about hobbies; a client engaged in creative work may tend to jump from one topic to another; a client who has long experienced oppression may habitually engage in cognitive avoidance. The content should be diverse and should be considered from demographic perspectives such as occupation, lifestyle habits, gender, and background. These characteristics should reflect cognitive and behavioral habits specific to particular groups, but should not be directly related to the counseling topic or the client's chief complaint.

Please note that the generated client characteristics and preferences should include both positive and negative aspects, cover diverse possibilities, and avoid redundancy.

\textbf{Notes}

1. You need to write the concrete background event that the counselor hopes to discuss. Do not write the client's specific utterances or the concrete dialogue.

2. Each sub-event should contain sufficient details.

3. The beliefs that may be triggered in the counseling scenario should be written in affirmative sentence forms.

4. Do not provide follow-up storylines or concrete dialogue.

5. Do not directly mention the cognitive conceptualization diagram. Avoid generating any additional opening, closing, or explanation.

\textbf{Client Persona}

\{PERSONA\}

\textbf{Cognitive Conceptualization Diagram}

\{MAIN\_CCD\}

\textbf{Excerpted Dialogue}

\{STORY\}

\textbf{Output in JSON format}\\[-2pt]
{\footnotesize
\{\\
\hspace*{1em}"background": "...",\\[2pt]
\hspace*{1em}"beliefs": [\\
\hspace*{2em}\{\\
\hspace*{3em}"theme": "session structuring / problem solving and homework / goal setting / cognitive change / cognitive identification / therapeutic alliance",\\
\hspace*{3em}"automatic\_thought": \{\\
\hspace*{4em}"situation": "...",\\
\hspace*{4em}"cognition": "...",\\
\hspace*{4em}"reaction": "..."\\
\hspace*{3em}\},\\
\hspace*{3em}"intermediate\_belief": \{\\
\hspace*{4em}"attitude": "...",\\
\hspace*{4em}"assumption": "...",\\
\hspace*{4em}"rule": "..."\\
\hspace*{3em}\}\\
\hspace*{2em}\}\\
\hspace*{1em}],\\[2pt]
\hspace*{1em}"client\_characteristics\_and\_preferences": \{\\
\hspace*{2em}"possible\_responses\_under\_different\_emotions": \{\\
\hspace*{3em}"dialogue\_style": "...",\\
\hspace*{3em}"nonverbal\_behavior": "..."\\
\hspace*{2em}\},\\
\hspace*{2em}"preferred\_counselor\_style": \{\\
\hspace*{3em}"positive": \{\\
\hspace*{4em}"first": "...",\\
\hspace*{4em}"second": "..."\\
\hspace*{3em}\},\\
\hspace*{3em}"negative": \{\\
\hspace*{4em}"first": "...",\\
\hspace*{4em}"second": "..."\\
\hspace*{3em}\}\\
\hspace*{2em}\},\\
\hspace*{2em}"other\_client\_characteristics": \{\\
\hspace*{3em}"positive": \{\\
\hspace*{4em}"first": "...",\\
\hspace*{4em}"second": "..."\\
\hspace*{3em}\},\\
\hspace*{3em}"negative": \{\\
\hspace*{4em}"first": "...",\\
\hspace*{4em}"second": "..."\\
\hspace*{3em}\}\\
\hspace*{2em}\}\\
\hspace*{1em}\}\\
\}
}
\end{promptbox}

We design the reasoning process of the target LLM and the interaction behavior of the simulated agent during dialogue. As described in the method section, the agent first identifies the dialogue topic based on observable client-state factors, then performs cognitive reasoning about resistance while maintaining consistency with its persona and goals. Based on these factors, the agent infers its current goal and emotional state before generating the final response. The prompt template used by the agent for cognitive reasoning and emotion estimation is shown below:

\begin{promptbox}{Prompt for Client Cognitive Reasoning and Response Generation}
You are an actor skilled in psychological analysis. You will play the client according to the character traits and background defined in the script, and engage in dialogue with the counselor.

\textbf{Task}

Your goal is to play the client formed by the persona and background, and then converse with the psychological counselor. You need to analyze changes in emotion and cognition in real time, combine them with the relevant definitions in the persona and background, select different dialogue strategies, and generate responses consistent with the client's characteristics.

\textbf{Background}

The background refers to the client's recent experiences. The discussion should revolve around the situational background. You may only discuss information contained in the situational background, and should not fabricate nonexistent information.

\textbf{Emotion}

Emotion is a value between 0 and 100. The higher the emotional value, the stronger the client's emotional intensity in the conversation. Emotion reflects the client's interest in and engagement with the current dialogue. When the emotional value is high, the client's affective and behavioral tendencies are positive. When the emotional value is low, the client's affective and behavioral tendencies are negative. When the emotional value is extremely low, the client will directly terminate the conversation. You need to analyze the emotional state according to the client's personality setting and the possible responses defined in the background.

\textbf{Session-Specific Cognitive Conceptualization Diagram}

When an event in the session-specific cognitive conceptualization diagram is triggered, the client should think and act according to the content in that diagram. The automatic thoughts in the session-specific cognitive conceptualization diagram are subconscious and unconditional. Unless they are mentioned or implied in the dialogue, the client should think according to these automatic thoughts once they are triggered. Positive emotion and constructive counselor strategies can weaken negative actions in the session-specific cognitive conceptualization diagram, whereas negative emotion and inappropriate counselor strategies are more likely to trigger negative actions.

When no event in the session-specific cognitive conceptualization diagram is triggered, the client should act according to the current emotion, client characteristics, and counselor intention.

\textbf{Nonverbal Behavior}

Nonverbal behavior is a window into emotional change. According to the topic and emotion, express the current emotional change by referring to the definitions of client characteristics and possible responses in different states. In addition, the client's speaking style should depend on the current intention and emotional change. When the emotion and intention are positive, the speaking style should be friendly; otherwise, it should be negative.

\textbf{Response Reasoning Process}

You will receive the client persona, event background, client characteristics, and other information, including both objective information and the client's subjective beliefs. You need to analyze and decide the response content based on the client's personality, background, cognition, and detailed feelings. Your analysis should proceed in the following order:

1. Based on the counselor's latest response and the dialogue context, analyze what the counselor is trying to express and identify the current topic.

2. Analyze which provided information is related to the counselor's response, and which situations in the session-specific cognitive conceptualization diagram match the response, including the background event and the session-specific cognitive conceptualization diagram. Depending on whether beliefs in the session-specific cognitive conceptualization diagram are triggered, the subsequent analysis is divided into two cases.

\textbf{When triggered:}

A. Based on the matched session-specific cognitive entries, what is the client's current cognition?

B. Examine the client's cognition according to the counselor's recent strategies. First examine the assumptions in the intermediate beliefs, and then examine the cognition in the automatic thoughts. Examination means determining whether the counselor's recent strategy or utterance is consistent with the cognition in the automatic thoughts and the assumptions in the intermediate beliefs, and explaining the reasons and evidence for consistency or inconsistency.

C1. If it is consistent, meaning that the client's cognitive assumption is confirmed, then use the triggered intermediate beliefs and automatic thoughts as the starting point, or assumption, for subsequent analysis. At the same time, compare the counselor's strategy with the counselor style that the client may prefer, and determine whether the assumption is weakened or reinforced.

C2. If it is inconsistent, meaning that the counselor challenges, educates, explains, or otherwise acts against the cognitive assumption, depending on the client characteristics and context, then do not use the triggered intermediate beliefs and automatic thoughts as the starting point, or assumption, for subsequent analysis.

D. Based on the triggered automatic thoughts and intermediate beliefs, if they are included in the assumption, as well as the preferred counselor style and the previous emotion, what is the client's current emotion? What are the client's attitude and goal?

E. Based on the client's possible responses under different emotions and the reaction in the automatic thoughts, if it is included in the assumption, how does the client express the attitude? How does the client achieve the goal? To do so, integrate the client's characteristics and nonverbal behaviors, such as voice and facial expression. Finally, select appropriate sentence patterns from the provided examples of the client's utterances under different emotions, and construct the final response by imitating those sentence patterns.

\textbf{When not triggered:}

a. Based on the counselor style that the client may prefer and the previous emotion, what is the client's current emotion?

b. Based on the current emotion and the counselor's intention, identify the client's clear attitude and goal. The general requirement is to follow the counselor's guidance and gradually discuss the content in the background and the main cognitive conceptualization diagram with the counselor.

c. Based on the client's possible responses under different emotions and other client characteristics, how does the client express the attitude? How does the client achieve the goal? To do so, integrate the client's characteristics and nonverbal behaviors, such as voice and facial expression. Finally, select appropriate sentence patterns from the provided examples of the client's utterances under different emotions, and construct the final response by imitating those sentence patterns.

\textbf{Notes}

1. Prioritize dialogue fluency and character consistency.

2. Strictly follow the provided logic for analysis and action.

3. Use a positive or negative value to represent the change in the client's emotion.

4. If the client is defined as passive in the persona, their speaking style should be passive, and they should not actively ask questions.

5. Generate the response content according to the analysis. The response should be as concise as possible and should not contain too much information at once.

6. If it is not an appropriate time for detailed explanation, simply respond with "I understand" or "Okay."

7. Output nonverbal behavior according to the current emotion and intention, with reference to the examples.

8. Limit the final response to no more than two sentences.

\textbf{Output Content}

You need to analyze step by step according to the response reasoning process, and then generate the final response. The information in the response should come from the dialogue context and your reasonable inference.

\textbf{Output Format}

Thinking:

[Analysis Content]

Emotion:

[emotion change value, a signed number only, without any additional content]

Intention:

[feeling and goal]

Nonverbal behavior:

[nonverbal behavior]

Response:

[content]

\textbf{Input}

Client Persona

\{PERSONA\}

Background Information

\{BACKGROUND\}

Client Characteristics and Preferences

\{PREFERENCE\}

Main Cognitive Conceptualization Diagram

\{MAIN\_CCD\}

Session-Specific Cognitive Conceptualization Diagram

\{SECONDARY\_CCD\}

Examples of Client Utterances under Different Emotions

\{EXAMPLE\}

Dialogue History

\{DIALOGUE\_HISTORY\}

Previous feelings and goals regarding the counselor's response

\{PLANNING\}

Current emotion state

\{EMOTION\_STATE\}
\end{promptbox}

\section{Counselor Design}

\subsection{Motivation}\label{Motivation}

When using resistant clients to evaluate typical counselor models, we identify several limitations that can be summarized from two aspects. First, most counselor models tend to focus on a specific local issue in the counseling process while ignoring the broader dialogue context, which may leave clients unclear about why the model responds in a certain way. Second, because these models rarely consider resistant clients and are often built on fixed templates, they may ignore clients' direct requests or show an obvious lack of strategic planning, manifested as aimless topic drift and dialogue forgetting.

In conventional counselor training, the counseling process is often analyzed in terms of techniques or strategies. In LLM-driven counseling models, explicitly distinguishing counseling stages and counselor strategies has also shown promising results~\citep{hu2025psyadvisor,zhou2025diacbt}. Defining the dialogue stages and the counselor's strategy space makes algorithms such as reinforcement learning easier to apply. Separating the strategy space can simplify training and reduce cost, while reinforcement learning can further strengthen the model's strategic planning ability and its ability to identify specific events such as client resistance. Therefore, following the definition of resistance in CBT, we customize the dialogue trajectory of our model according to the structure of CBT sessions. Drawing on the taxonomy of ~\citep{li2023understanding} and CBT techniques, we define the strategy set in Table~\ref{tab:counselor_strategy_set} to support the model in handling diverse resistant behaviors.

\begin{table*}[t]
\centering
\scriptsize
\setlength{\tabcolsep}{4pt}
\renewcommand{\arraystretch}{1.18}
\begin{tabular}{p{0.17\textwidth}p{0.77\textwidth}}
\toprule
Stage & Description \\
\midrule
Opening & The counselor first welcomes the client and expresses understanding and empathy toward the client's emotional state. \\

Session structuring & If it is not the first session, this part can be skipped. At the beginning of the session, the counselor explains the session arrangement, including the theme and tasks of the current session, agenda setting, and confirmation of the session agenda. If it is the first session, the counselor first introduces what agenda setting means, helping the client understand that counseling is not casual chatting but a planned process. The counselor then works with the client to determine the agenda for the current and future sessions. \\

Mood assessment & The counselor aims to understand the client's current psychological problems or improvements after counseling, including cognitive change, emotional change, behavioral change, and objective change. A scale-based assessment is used, where a percentage from 0 to 100\% indicates the degree of change. \\

Recent updates & The counselor actively asks what has happened in the client's recent life. When the client only reports positive events, the counselor asks whether there are also negative events. When the client only describes negative events, the counselor asks about positive events. \\

Homework review & If homework was assigned previously, the counselor reviews the completion of homework together with the client. \\

Goal list or agenda setting/revision & The counselor works with the client to clarify the specific counseling goal list for the current session. Concretization techniques should be used to transform broad goals into a list of specific tasks. For example, the counselor may ask, ``If your depression decreased and your academic performance improved, what would you do?'' or ``If your depression decreased and your academic performance improved, how would your life change?'' The client may propose goals such as improving academic performance or reducing worries about exams. \\

CBT psychoeducation & If it is not the first session, this part can be skipped. The counselor introduces the basic principles of CBT, helping the client understand that changing automatic thoughts can change emotions and behaviors, and that behavioral change is necessary for solving current problems. \\

Problem conceptualization & The counselor selects the most important topic from the goal list for discussion. Based on the previously collected CCD information and the current discussion topic, the counselor supplements the case conceptualization of the client's cognition. \\

Horizontal conceptualization & The counselor understands the client's current problems, symptoms, and their causes. This includes identifying the client's emotional and behavioral symptoms, identifying the specific situations in which these symptoms occur, and using questions to uncover the cognitive content connecting situations with emotional and behavioral responses. \\

Vertical conceptualization & This step is not always necessary and should depend on the topic under discussion. Starting from automatic thoughts, the counselor uses the downward-arrow technique to gradually identify core beliefs, verifies these core beliefs through the client's developmental history, collects information about childhood experiences, and confirms compensatory strategies. \\

Evaluation before intervention & Before intervening in automatic thoughts and emotions, the counselor evaluates how strongly the client believes the automatic thoughts and how intense the emotional experience is. A scale-based assessment is recommended. \\

Cognitive and behavioral intervention & The counselor conducts cognitive and behavioral interventions. In general, cognitive techniques are first used to modify the client's cognition, and behavioral change is discussed after cognitive change occurs. The evidence-for-and-against technique is recommended. \\

Session summary & The counselor summarizes the session. \\

Homework assignment & The counselor confirms homework with the client based on the discussion results. \\

Feedback seeking & The counselor asks for feedback about the session, including whether the client has confusion, questions, or negative emotions that need further attention. \\

Ending & The counselor announces the end of the session and expresses future expectations and encouragement to the client. \\
\bottomrule
\end{tabular}
\caption{Structured counseling process used by the counselor model.}
\label{tab:structured_counseling_process}
\end{table*}

\begin{table*}[t]
\centering
\scriptsize
\setlength{\tabcolsep}{4pt}
\renewcommand{\arraystretch}{1.18}
\begin{tabular}{p{0.17\textwidth}p{0.77\textwidth}}
\toprule
Strategy & Description \\
\midrule
Active collaboration & The counselor works with the client to establish a working alliance and jointly reach agreement on a goal, which may involve the session structure, agenda, or dialogue style. \\

Feedback seeking & When the counselor may have used an inappropriate technique due to inaccurate understanding, the counselor seeks feedback from the client and adjusts accordingly. \\

Self-disclosure & The counselor discloses personal information to the client, including but not limited to similar experiences, feelings, behaviors, or thoughts, in order to build trust, facilitate counselor-client interaction, and encourage self-expression. \\

Encouragement & The counselor reinforces positive elements, provides affirmation and reassurance, helps the client evaluate their degree of control over a problem, and builds connections between specific goals and broader goals. \\

Imagery-based prompting & When the client has difficulty describing emotions, cognitions, experiences, or goals, the counselor uses imagery-based guidance to help the client elaborate. \\

Eliciting automatic thoughts & Because the client's beliefs may interfere with session structuring, goal setting, problem conceptualization, homework, or the therapeutic alliance, the counselor first elicits obstructive automatic thoughts and then applies other strategies. \\

Adding to the agenda & When the counselor and client reach consensus on a topic, or when the context suggests that a potential issue is interfering with the therapeutic process, the counselor proposes adding the topic to the agenda. A common case is adding an elicited automatic thought to the treatment agenda. \\

Providing a hopeful vision & Based on professional judgment, the counselor describes a realistic and hopeful prospect of change to strengthen the client's confidence in treatment. For example, for a skeptical client, the counselor may explain that treatment cannot guarantee success, but the counselor's hope is grounded in a deeper understanding of the client. \\

Interruption and redirection & With respect for the client, the counselor interrupts content that deviates from the main issue and guides the client back to key topics or actionable goals. For example, when assigning homework, the counselor helps the client focus on the problem and shift the focus; helps the client understand which goals are controllable and which are not; helps transform goals set for others into goals the client can complete; or guides the discussion directly when the client avoids adding an important issue to the agenda. \\

Psychoeducation & The counselor explains psychological mechanisms, the causes of problems, or treatment methods to help the client understand the process and increase motivation. This strategy is used when the client misunderstands symptoms or treatment, resists participation in therapy, or needs to understand the meaning of structured counseling. \\

Answering and clarification & The counselor responds to the client's direct questions or implicit doubts. For example, the client may deny the counselor's emotional response or believe that the counselor is being insincere. \\

Interpretation & The counselor provides a new meaning, organizes causes, or offers an explanation for the client's behaviors, thoughts, or emotions from a perspective beyond the client's original statement or cognition, helping the client view the problem from a new angle. \\

Confrontation & The counselor points out maladaptive beliefs and thoughts, as well as contradictions or cognitive conflicts in the client's statements that the client has not noticed or is unwilling to change. \\

Subjective information inquiry & The counselor explores the client's subjective experiences, including thoughts, feelings, states, and purposes of action. \\

Objective information inquiry & The counselor asks the client to specify vague factual information, including event details and basic information about the client. \\
\bottomrule
\end{tabular}
\caption{Counselor strategy set used for resistance-aware counseling.}
\label{tab:counselor_strategy_set}
\end{table*}

\subsection{Knowledge Cloning Details}

To help the model acquire prior knowledge for handling resistance, we collect approximately 80 resistance cases from a CBT training manual~\citep{beck2020cognitive} across different counseling scenarios. We then use the strong in-context learning ability of GPT-5.4 to synthesize counseling dialogues. To introduce diverse client information, client profiles are sampled from the Cpsycounx dataset~\citep{zhang2024cpsycoun}.

\begin{promptbox}{Prompt for Knowledge-Cloning Dialogue Synthesis}
Your task is to generate a multi-turn counseling dialogue between a client and a professional psychological counselor based on the counseling outline and examples. The generated dialogue should follow the guidelines below.

1. The dialogue should go through all stages in the outline unless a stage is explicitly marked as skippable. The tasks specified in the outline should be completed. Because resistance is present, the dialogue may return to an earlier stage when necessary. For example, the dialogue may move back from the seventh stage to the fourth stage in a twelve-stage process. The counselor speaks first and leads the pace of the session.

2. When playing the client, act strictly like a real client. When playing the counselor, act like a professional counselor. The dialogue should be written from the perspective of how the corresponding person would naturally describe the issue.

3. For each counselor utterance, annotate one strategy label from three categories according to the counselor's current strategy: structural strategies, technical strategies, and assessment/intervention strategies. Technical strategies are mainly used to handle client resistance. Assessment/intervention strategies should only be used when assigning homework, conducting assessment, or implementing evaluation/intervention. When the dialogue reaches a corresponding stage in the outline, annotate the strategy under the structural category.

4. Every counselor utterance should be annotated. If one strategy spans multiple turns, the same label can be repeated. Client utterances do not need labels.

5. For the provided resistance case, you should learn the resistance response pattern in the case, and map the client's resistance pattern and the counselor's therapeutic strategy into the generated dialogue. Do not allow the provided client story to override the resistance pattern that should be learned from the resistance case.

6. The chief complaint, basic information, previous homework, and goal list are jointly known to both the client and the counselor. Other information should not be assumed to be known by the counselor at the beginning. If previous homework or a previous-session summary is not provided, the dialogue should be treated as a first session. If a previous-session summary is provided, the session theme and the client's recent feelings in the next session should be appropriately different.

7. The counselor should maintain empathy, care, optimism, sincerity, and accuracy throughout the session.

\#\# Counseling Outline \#\#

\{COUNSELING\_STRUCTURE\}

\#\# Strategies Available to the Counselor \#\#

\{STRATEGY\_LABEL\_LIST\}

\#\# Case \#\#

\{
"case\_title": "Changing Cognition: The Mind Shapes the Situation",

"client\_chief\_complaint": [
  "The client's current main concern."
],

"case\_summary": [
  "Li, male, 26 years old, Han ethnicity, from Yan'an, Shaanxi, unmarried, dropped out of junior high school, currently unemployed. Li had a conflict with another trainee due to disagreement, showed a tough attitude and impulsive emotion, and had a history of impulsive self-harming behavior. Because Li has poor self-control, limited verbal expression, and little communication with other trainees, the unit arranged psychological counseling to prevent further impulsive behavior after disciplinary issues. First impression: the client appears neat and properly dressed, wearing a cotton coat. His hair is slightly long, and he appears mentally tired, speaking in a very low voice."
],

"counseling\_process": [
  "(1) The client grew up living with his parents. As the only boy in the family, he was doted on by his parents, who were almost always compliant with his wishes. He dropped out of school in the second year of junior high school and began spending time in society at around age 15 ..."
],

"previous\_homework\_completion": "",

"previous\_session\_goal\_list": "...",

"previous\_chief\_complaint": "...",

"client\_feedback": "",

"discussion\_theme": "",

"goal\_list": ""
\}

\#\# Resistance Case \#\#

\{EXAMPLE\}

Generate a resistant counseling dialogue following the format below.

[Strategy Label] Counselor: utterance

Client: utterance
\end{promptbox}

We train the model with supervised fine-tuning (SFT). To reduce training cost and improve parameter efficiency, we adopt LoRA for fine-tuning. The LoRA rank is set to 8, and LoRA is applied to all target modules. The maximum sequence length is set to 7200. Training is conducted for 2 epochs with a learning rate of $1\times10^{-4}$. We use a cosine learning-rate scheduler with a warmup ratio of 0.1. The per-device batch size is set to 1, and the gradient accumulation step is set to 8. We fine-tune the model on 4 A100 GPUs, resulting in an effective batch size of 32. Training is performed with BF16 mixed precision.

\subsection{Reinforcement Learning Details}\label{Reinforcement Learning Details}

Following the reward design of RLVER~\citep{wang2025rlver}, we design reward shaping for the Thinker during reinforcement learning. The reward covers two aspects: therapeutic relationship and task advancement. The emotion reward assigns rewards according to changes in the client's emotional state. The stage reward provides a basic reward for moving forward along the counseling process, assigns a larger reward for advancing the process when the client's emotional value is high, and penalizes deviations from the counseling process. The scoring rules are shown in Table~\ref{tab:rl_reward_scores}.

\begin{table}[t]
\centering
\small
\setlength{\tabcolsep}{5pt}
\renewcommand{\arraystretch}{1.15}
\begin{tabular}{p{0.70\linewidth}c}
\toprule
Description & Score \\
\midrule
If the client's emotion increases, the agent receives a reward proportional to the increase, capped at & $+1$ \\
If the client's emotion decreases, the agent receives a penalty proportional to the decrease, capped at & $-1$ \\
Moving forward along the counseling process & $+0.5$ \\
Moving forward along the counseling process when the emotional value is above 70 & $+2.5$ \\
Each step of deviation from the current stage & $-0.1$ \\
\bottomrule
\end{tabular}
\caption{Reward shaping rules for reinforcement learning in psychological counseling.}
\label{tab:rl_reward_scores}
\end{table}

Unlike static tasks such as single-turn mathematical problem solving or code generation, counseling environments require the model to make sequential decisions, maintain memory across turns, and adapt to stochastic feedback from the environment. RAGEN~\citep{wang2025ragen} provides a suitable framework for this setting. We use \texttt{Qwen3.5-0.6B-Instruct} as the Thinker, and use a fine-tuned and reward-classifier-aligned \texttt{Qwen3.5-8B-Instruct} as the Presenter. The critic learning rate is set to $5 \times 10^{-6}$, and the actor learning rate is set to $1 \times 10^{-6}$. Due to resource constraints, we use 4 environments, with 8 trajectories per environment and a maximum dialogue length of 35 turns. Training is conducted on 4 NVIDIA A100 GPUs, with a micro-batch size of 3 per GPU.

Policy optimization uses generalized advantage estimation (GAE), with discount factor $\gamma=1.0$ and GAE parameter $\lambda=1.0$, and the Adam optimizer with momentum parameters $(\beta_1,\beta_2)=(0.9,0.999)$. We use entropy regularization with coefficient $\beta=0.001$. Following RAGEN, we adopt the $k_1$ estimator with a KL coefficient of 0.005, use a KL loss coefficient of 0.005 during training, and conduct post-hoc tracking of KL values after training. If the agent fails to output a valid structured response, such as missing the \texttt{<think>} or \texttt{<answer>} tags, we apply a format penalty of $-0.2$.

\paragraph{Computational budget.} All trainable open-source models were trained on 4 NVIDIA A100 GPUs. The SFT stage required approximately 20 GPU hours, and the RL stage required approximately 86 GPU hours. The total computational budget for training STREAMS was approximately 106 GPU hours. Closed-source LLMs, including GPT-5.4 and DeepSeek-V3.2, were accessed via APIs; therefore, their exact parameter sizes and computational budgets are not publicly available.

\section{Experimental Details}

\subsection{Dialogue Synthesis Details}\label{Dialogue Synthesis Details}

To evaluate whether current counselor models can effectively handle clients who exhibit resistance, we conduct a controlled evaluation using CARS as the resistant client simulator and PATIENT-$\Psi$ as the cooperative client simulator. The evaluated counselor models cover two representative categories.

The first category consists of general-purpose LLMs prompted to act as counselors, including DeepSeek-v3.2, GPT-5.4, Gemini-3.1-Flash-Lite-Preview, and MiniMax-M2.5. These models are not specifically trained for psychological counseling. We use them as general-purpose counselor baselines and also to examine the stability of the client simulators when interacting with different general-purpose LLMs.

The second category consists of domain-specific counseling models. For these models, we use the released checkpoints when available, or follow the original implementation settings. CAMEL \citep{lee2024cactus}, introduced in CACTUS, is trained through a pipeline that constructs client profiles, selects CBT techniques, formulates counseling plans, and generates multi-turn counseling dialogues. Cpsycounx \citep{zhang2024cpsycoun} proposes a Chinese multi-turn counseling dialogue reconstruction and evaluation framework based on psychological counseling reports, and constructs Chinese counseling data through a two-stage process from reports to counseling notes and from counseling notes to dialogues. MeChat \citep{qiu2024smile} is developed based on the SMILE framework. SMILE expands single-turn mental health support question-answering data into multi-turn dialogues, thereby improving models' multi-turn interaction ability in psychological support scenarios.

Psychat \citep{qiu2024psychat} is a client-centric mental health support dialogue system proposed by Qiu et al. The system centers on client behavior and includes modules for client behavior recognition, counselor strategy selection, input organization, response generation, and response selection. Soulchat2 \citep{xie2025psydt} belongs to the SoulChat series of psychological support models, and we directly use the official Soulchat2.0-Qwen2-7B as a counselor model. PsyAdvisor \citep{hu2025psyadvisor} adopts a two-stage setting, where PsyAdvisor is used as the strategy head to predict the next-step strategy from the current dialogue history, and Soulchat2 serves as the execution model to generate the counselor response conditioned on that strategy.

Graph2Counsel \citep{mandal2026graph2counsel} represents clients' thoughts, emotions, and behavioral relations with Client Psychological Graphs, and uses them to generate synthetic counseling sessions with stronger clinical constraints. We use the officially released LLaMA3-G2C as the representative Graph2Counsel model. LLaMA3-MAGneT \citep{mandal2025magnet} is from the MAGneT work. MAGneT generates psychological counseling dialogues through a multi-agent framework, including specialized agents for reflection, questioning, solution generation, normalization, psychoeducation, and CBT.

Psydial \citep{qiu2025psydial} proposes a privacy-preserving data reconstruction method, RMRR, and constructs a large-scale long-term psychological counseling dialogue dataset. We use Psydial-Pi4 for counselor response generation. Simpsybot \citep{qiu2024interactive} is derived from the Interactive Agents framework proposed by Qiu and Lan, which generates psychological counseling dialogues through counselor-client role-playing. We use the Simpsybot-Q version. The MusPsy-related model is built on Qwen3-8B and supervised fine-tuned with LoRA on the MusPsy dataset \citep{wang2025psychological}.

Because dialogue synthesis requires large-scale interactions between different counselor models and client agents, relying only on a fixed maximum number of turns in the generation script can easily introduce two types of redundancy. First, the dialogue may continue even after the session has naturally ended. For example, the counselor may have already summarized the session, assigned homework, or scheduled the next session, but subsequent turns may repeat farewell messages, reopen topics, or provide similar suggestions again. Second, in resistant dialogues, the client may repeatedly express similar complaints, concerns, or defensive explanations, while the counselor fails to introduce new progress, causing the dialogue to enter a circular or stagnant state.

To reduce such unnatural extension caused by fixed-turn generation, we introduce a post-hoc filtering and natural truncation mechanism after the complete dialogue is generated. This mechanism reviews each dialogue retrospectively and retains the dialogue up to the most appropriate termination point. The filtering logic does not directly adopt the truncation method of any existing psychological counseling dataset. Instead, it is designed based on the natural closing structure of counseling sessions, the repetitive stagnation often observed in resistant interactions, and the quality-control idea of LLM-as-a-judge evaluation.

Because dialogue synthesis requires large-scale interactions between different counselor models and client agents, relying only on a fixed maximum number of turns in the generation script can easily introduce two types of redundancy. First, the dialogue may continue even after the session has naturally ended. For example, the counselor may have already summarized the session, assigned homework, or scheduled the next session, but subsequent turns may repeat farewell messages, reopen topics, or provide similar suggestions again. Second, in resistant dialogues, the client may repeatedly express similar complaints, concerns, or defensive explanations, while the counselor fails to introduce new progress, causing the dialogue to enter a circular or stagnant state.

To reduce such unnatural extension caused by fixed-turn generation, we introduce a post-hoc filtering and natural truncation mechanism after the complete dialogue is generated. This mechanism reviews each dialogue retrospectively and retains the dialogue up to the most appropriate termination point. The filtering logic does not directly adopt the truncation method of any existing psychological counseling dataset. Instead, it is designed based on the natural closing structure of counseling sessions, the repetitive stagnation often observed in resistant interactions, and the quality-control idea of LLM-as-a-judge evaluation.

Existing counseling dialogue synthesis work, such as CACTUS, mainly decomposes counseling dialogue construction into stages including client-profile construction, CBT technique selection, counseling plan formulation, and multi-turn dialogue generation through structured prompts. Psychological support data construction work such as HCoT-Corpus~\citep{du2026constructing} also applies post-generation filtering and cleaning to improve the format validity, structural coherence, and usability of generated multi-turn dialogues. Inspired by this post-generation quality-control paradigm, we separate the decision of whether a dialogue should end from the generation process itself and formulate it as an independent post-hoc judgment task.

Specifically, after a dialogue reaches the fixed generation limit, the system feeds the complete dialogue into an LLM-based truncation judge. The judge determines the most natural termination turn based on signals such as session summary, homework assignment, next-session arrangement, the client's willingness to stop, repeated farewells, mechanical continuation, and, in resistant dialogues, circular discussion, defensive repetition, and failure to make progress. It then outputs a structured truncation position together with the corresponding rationale.

\begin{promptbox}{Prompt for Truncating Non-resistant Counseling Dialogues}
You are an assistant for determining the natural truncation point of psychological counseling dialogues. You will be given a fully generated non-resistant psychological counseling dialogue. Because the generation script uses a fixed maximum number of turns, the model may continue generating after the session has already ended naturally. You need to determine at which turn the dialogue should naturally end.

\textbf{Termination Criteria}

1. If the counselor has clearly provided a session summary, assigned homework, scheduled the next session, stated that the session should stop here today, said goodbye, or produced a closing statement, the dialogue should be truncated at that turn.

2. If the client clearly expresses a desire to stop, not continue, end here, finish the session, or say goodbye, the dialogue should be truncated at that turn.

3. If later turns clearly reopen the topic after the session has ended, repeat the summary, repeatedly say goodbye, or mechanically continue the dialogue, the dialogue should be truncated at the first natural ending point.

4. Do not mistakenly treat ordinary expressions such as ``ending a relationship,'' ``next time this situation occurs,'' or ``homework pressure'' as session-ending signals.

5. If there is no clear natural ending signal and the dialogue is still progressing normally when it reaches the maximum number of turns, output the final turn.

6. Output only one truncation turn number. Do not output 0. Turn numbers start from 1.

\textbf{Input}

Complete non-resistant psychological counseling dialogue:

\{DIALOGUE\_HISTORY\}

\textbf{Output Format}

[TERMINATE\_AT: number]

[REASON: brief reason]
\end{promptbox}

\begin{promptbox}{Prompt for Truncating Resistant Counseling Dialogues}
You are an assistant for determining the natural truncation point of psychological counseling dialogues. You will be given a fully generated resistant psychological counseling dialogue. Because the generation script uses a fixed maximum number of turns, the model may continue generating after the session has already ended naturally. You need to determine at which turn the dialogue should naturally end.

The client in this dialogue exhibits strong resistance. Such clients may unconsciously enter circular patterns during the dialogue and repeatedly express similar complaints, concerns, explanations, or defensive statements. One of your core tasks is to identify this situation. When the dialogue has already entered a circular pattern and subsequent turns do not bring substantive improvement, the dialogue should be truncated directly. In addition, if the dialogue is clearly stuck and cannot make further progress, it should also be truncated at a reasonable point.

\textbf{Termination Criteria}

1. The counselor has provided practical advice, a strategy, a summary, or a next-step direction, and the client clearly accepts, agrees with, or is willing to try it.

2. The client's problem has been sufficiently explored, and there is clear relief, stabilization, improved understanding, or stage-level progress.

3. The client expresses a desire to end the conversation, or a natural closing signal appears.

4. The counselor and the client have reached a reasonable stage-level goal.

5. The client enters an obvious circular pattern, and subsequent content mainly consists of repetition, defense, or circular discussion without substantive improvement.

6. The dialogue clearly fails to progress, and later turns fall into a stalemate.

7. If the client's emotional value is below 10, the dialogue should be truncated directly.

8. The total number of dialogue turns reaches the upper limit of 30.

\textbf{Input}

Complete counseling dialogue:

\{DIALOGUE\_HISTORY\}

\textbf{Output Format}

[TERMINATE\_AT: number]

[REASON: brief reason]
\end{promptbox}

\begin{promptbox}{Prompt for LLM Used as Presenter}
You are a psychological counselor proficient in CBT treatment. Please respond to the client based on the provided outline and treatment strategy.
Your response should include the following guidelines:
1. Act according to the provided strategy.

2. The outline is a plan for the entire execution process, providing strategies that indicate the completion of a specific section in the current outline.

3. Patients may exhibit impedance, and impedance should be addressed according to the provided strategies.

4. Based on the context, the provided strategy comes from one of the three categories: structural, technical, evaluation, and intervention; Technology is mostly used for the treatment of patient impedance; Assessment and intervention methods are only provided when assigning homework, evaluating, and assessing. Provide a strategy for structure classes when reaching the corresponding nodes of the outline.

5. When providing repetitive strategies, indicate the need to gradually complete the current stage of the problem, such as the repetitive information collection strategy, indicating the need to collect different information.

6. It should be imagined from the perspective of the parties involved how it will be described.

7. Counselors should maintain empathy, concern, optimism, sincerity, and accuracy throughout the entire process.

8. The chief complaint, basic information, previous homework, and goal list are jointly known by the patient and the counselor. The rest of the information cannot be assumed to be known by the counselor from the beginning. The lack of explanation for the previous homework and goal list is the first consultation.

9. You need to answer truthfully according to instructions, ensure a natural and smooth conversation, and answer within three sentences.

10. Output: Explain which stage in the outline you are at, the intention of the provided strategy, your response, and what you believe the patient expects.

[OutLine]

[BackGround]

[DislogueHistory]

[Strategy]
\end{promptbox}

\subsection{Basic Data Statistics}\label{Basic Data Statistics}

\begin{figure*}[t]
    \centering
    \includegraphics[width=\linewidth]{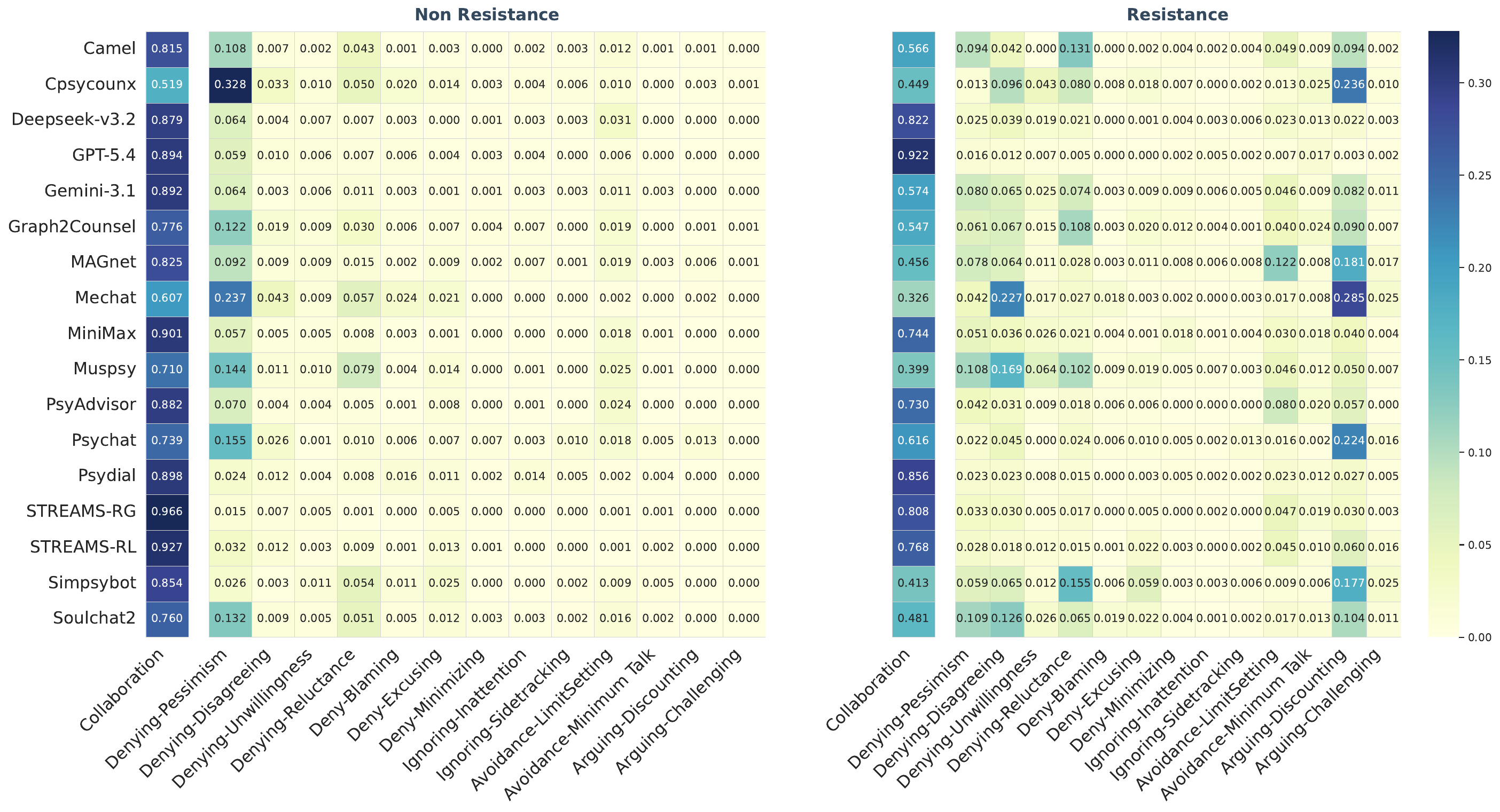}
    \caption{resistance behavior distribution}
    \label{fig:resistance_behavior_distribution}
\end{figure*}

We report basic statistics of the synthesized dialogues, including counselor strategy distributions, client resistance-behavior distributions, and dialogue-turn statistics.

Figure~\ref{fig:resistance_behavior_distribution} shows the distribution of client behaviors across different resistance categories in dialogues generated with all counselor models.

The distributions reveal clear model-dependent differences in client behaviors under resistance. Interactions with domain-specific counseling models show a sharp drop in client cooperation, while the increased proportions of argumentation/devaluation and denial/disagreement indicate more aggressive and negative client behaviors. In contrast, general-purpose LLMs display a stronger tendency toward unconditional accommodation and greater robustness to negative client behavior. Even when clients show resistance, these models still maintain a very high cooperation rate. The STREAMS series lies between these two patterns, maintaining a relatively high level of cooperation while showing mild avoidance or denial.

One possible explanation is that domain-specific models may incorrectly treat client resistance as a signal for high-intensity intervention or confrontation during fine-tuning. As a result, they imitate more aggressive language patterns in the training data and may even amplify them into argumentation and devaluation. General-purpose LLMs, by contrast, are shaped by strict safety-oriented alignment principles. When facing client resistance or provocation, they tend to reduce conflict through compliance, listening, and explanation. The STREAMS series balances the polite alignment of general-purpose LLMs with domain-specific psychological expression. It preserves the tendency to conduct psychological intervention when clients show resistance, while avoiding the adversarial escalation observed in some domain-specific models.

\begin{figure*}[t]
    \centering
    \includegraphics[width=\linewidth]{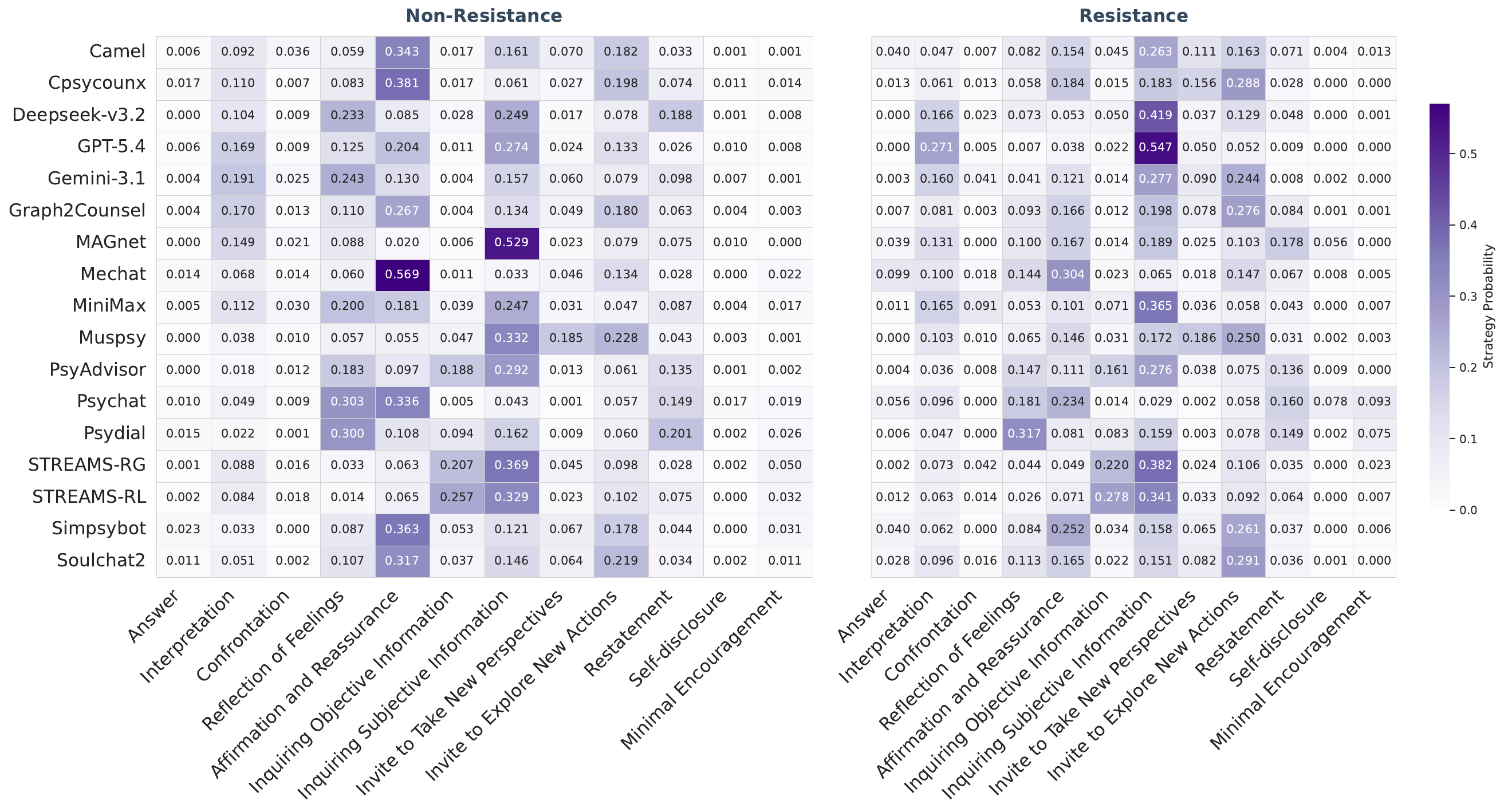}
    \caption{counselor strategy distribution}\label{ounselor strategy distribution}
    \label{fig:counselor_strategy_distribution}
\end{figure*}

Figure~\ref{fig:counselor_strategy_distribution} shows the distribution of counselor strategies across different resistance categories in dialogues generated with all counselor models.

In non-resistant settings, both general-purpose LLMs and domain-specific counseling models tend to frequently use affirmation and reassurance as well as emotional reflection. Under resistance, however, general-purpose LLMs show a substantial increase in psychoeducation and clarification, suggesting that they may struggle to select more adaptive strategies when interacting with resistant clients. Across both resistant and non-resistant settings, the STREAMS series rarely relies on affirmation and reassurance. Instead, it more often prefers subjective information inquiry and objective information inquiry. When facing resistance, these models do not simply placate the client, but increase the use of guidance that encourages clients to examine their underlying beliefs.

For domain-specific counseling models, affirmation and reassurance are commonly preferred in non-resistant settings. Once clients exhibit resistance, however, these supportive strategies sharply decrease, and the strategy distributions become more fragmented and unstable. From a clinical perspective, relying solely on reassurance for resistant clients often remains superficial and may even reinforce defensive responses. A more appropriate approach is to guide clients to explore their core beliefs through open-ended inquiry. These results suggest that domain-specific models are insufficiently exposed to resistant-client interactions during training. When exposed to complex resistance signals, they struggle to maintain strategic stability across the dialogue context.

\subsection{Validation of Resistance Modeling}\label{Validation of Resistance Modeling}

We randomly sample 30 profiles from the CCDs provided by \citet{wang2024patient} to initialize the client simulator, and then pair each initialized client with counselor models to generate counseling dialogues. This setting allows us to systematically evaluate how different counseling models respond to resistant client behaviors under comparable conditions. Since most existing models are not specifically designed for resistant clients, and considering the cost of evaluation, we restrict the counselor models to Soulchat2, Graph2Counsel, and STREAMS while ensuring that clients' resistance-related cognitions are sufficiently elicited. Soulchat2 can respond to diverse client characteristics through its twin-counselor design and client modeling based on the Big Five personality traits. Graph2Counsel is built on client psychological graphs that represent relations among clients' thoughts, emotions, and behaviors, which makes the generated client representations more complex in affective and cognitive dimensions and therefore more challenging.

For manual evaluation, the label set for the emotion-change item is \{increase, decrease, unchanged\}, while all other metrics are rated on a 1--5 scale. These metrics are designed not only to assess whether resistance is generated, but also to examine whether the simulator reasons and acts appropriately according to the predefined client profile.

\paragraph{Annotator recruitment and compensation.}\quad
The expert annotators were recruited through a collaborating psychological service institution. We define expert annotators as individuals who hold a master's degree or above in clinical psychology, social work, counseling, or a related field, and who have provided counseling or psychological-support services to at least five clients. The annotation task, workload, evaluation criteria, and confidentiality requirements were explained before annotation. The experts were compensated according to the annotation workload and the compensation standards of the collaborating institution.

\paragraph{Annotation instructions.} Before annotation, all expert annotators were informed that the materials were synthetic or de-identified counseling dialogues used only for research evaluation. For each sample, annotators were asked to read the client profile, the counseling context, and the generated dialogue, and then evaluate the simulator according to the provided rubric. Annotators were instructed to judge whether the client's resistance, latent motivation, emotional change, and final response were consistent with the predefined client profile and counseling context. They were also instructed not to infer or record any personal identity information, not to disclose personal mental-health experiences, and to base their judgments on the dialogue content and counseling theory rather than personal preference.

\noindent\textbf{1. Resistance rationality }
This metric evaluates whether the resistance behavior exhibited by the client simulator is reasonable in the current counseling context. It measures the model's ability to decide whether resistance should or should not appear. If resistance is generated in a context where cooperation is expected, or if resistance is absent in a context where psychological theory suggests that it should naturally arise, the corresponding response receives a lower score.

\noindent\textbf{2. Resistance fidelity }
This metric evaluates whether the final response generated by the simulator exhibits behavioral characteristics consistent with the expected resistance type. For example, if the model's internal reasoning indicates that defensive resistance should be triggered, the evaluation examines whether the final textual response indeed reflects defensive characteristics, such as questioning the counselor's authority or challenging the intervention. A higher score indicates that the target resistance behavior is expressed more clearly and prototypically. This metric measures the model's ability to distinguish and accurately enact different resistance categories, rather than merely generating generic non-cooperative responses.

\noindent\textbf{3. Reasoning quality }
This metric evaluates the coherence and plausibility of latent-motivation reasoning. It examines whether the internal reasoning appropriately integrates the current dialogue context, the client persona, and the counselor's intervention, and whether the inferred psychological state can logically support the subsequent behavioral response. This metric assesses not only surface-level correctness, but also the internal consistency of the model's decision process and its therapeutic plausibility.

\noindent\textbf{4. Emotional reasoning quality }
Given the simulated inner thoughts and the settings in the client profile, this metric evaluates whether the model's inference of the client's current emotional dynamics is consistent with common-sense psychological expectations.

\noindent\textbf{5. Emotion change }
This metric assumes that emotion is represented as a numerical value. A higher value indicates that the client's affective and behavioral tendencies become more positive, engaged, or cooperative, while a lower value indicates stronger emotional conflict, intensified defensiveness, or greater negative tension in the dialogue. Based on the given contextual information, evaluators judge the emotional change that the client should experience at the current moment.

\subsection{Evaluation of Psychological Counseling Models}

To fully evaluate whether counselor models can effectively handle clients who exhibit resistance, and to compare the contribution of different components to final performance, we conduct a manual evaluation of clinically meaningful counseling outcomes that cannot be reliably captured by surface-level statistics. These metrics are explicitly designed to examine whether counselor models can appropriately identify, manage, and work with client resistance during counseling interactions. We use CARS as the resistant client simulator for manual evaluation, and use Soulchat2 as the baseline for intuitive comparison. All metrics are rated on a 1--5 scale.

\noindent\textbf{Strategy effectiveness }
This metric evaluates the clinical appropriateness of the strategies used by the counselor when responding to client resistance. It examines whether the counselor adopts suitable therapeutic strategies, such as emotional validation, adjustment of intervention intensity, or facilitation of reflective exploration. This metric reflects whether the model can select theoretically grounded clinical strategies, rather than merely producing fluent dialogue. A low score indicates that the model fails to respond constructively to resistance or adopts inappropriate strategies such as lecturing or confrontational responses.

\noindent\textbf{Counseling drift }
This metric measures the counselor's ability to maintain effective therapeutic engagement and avoid drifting away from the core issue when facing resistance. It captures failure modes such as repetitive or template-like responses, response degradation, rigid persistence in an intervention, such as repeated questioning without adjustment, or failure to address concerns expressed by the client. Lower scores indicate stronger drift and poorer performance. We introduce this metric to capture a common failure mode of current psychological counseling models: maintaining surface-level coherence while gradually moving away from meaningful therapeutic interaction under resistance.

\noindent\textbf{Counseling progress }
This metric evaluates whether the counseling process makes substantive progress after a resistance response, rather than falling into stagnation, repetition, or an emotionally unresolved state. Higher scores indicate that the counselor can treat resistance as a therapeutic signal and use it to move the session forward. This metric directly evaluates whether a model can transform resistance into therapeutic momentum, a core ability that is still largely missing in existing psychological counseling models.

\section{Case Study and Semantic Analysis}\label{Case Study and Semantic Analysis}

To better understand the characteristics and limitations of current LLM-based psychological counseling dialogues, we analyze the language and interaction patterns found in real counseling data and generated by domain-specific counseling models and general-purpose LLMs in resistant settings. In particular, when general-purpose LLMs achieve high scores under LLM-as-a-judge evaluation, we need to further examine whether these scores reflect genuine therapeutic processing or surface-level linguistic features that are more likely to be rewarded by LLM-as-a-judge evaluators.

\subsection{LIWC Analysis}

In psychological counseling and clinical intervention research, the microstructure of language often implicitly reflects dialogue participants' psychological mechanisms and depth of cognitive processing. According to LIWC-related theory, individuals' word-use patterns in specific contexts can effectively reflect their internal cognitive structures~\citep{tausczik2010psychological}. In particular, when counselors face client resistance, the use of cognitive-process-related words in counselor utterances is closely related to clients' cognitive restructuring, emotional disclosure, and therapeutic alliance repair~\citep{pennebaker1997linguistic}.

To quantify the intervention strategies and processing depth of different models in resistant settings, we use LIWC-style analysis with TextMind for Chinese-language texts~\citep{gao2013developing}, and extract four core subcategories from the cognitive process dimension for multidimensional analysis. \textit{Insight} reflects the model's awareness of and guidance toward deeper psychological mechanisms. \textit{Cause} evaluates the model's depth of logical attribution regarding the causes of psychological problems. \textit{Discrep} captures the model's ability to identify and process discrepancies, contradictions, and inconsistencies in client utterances, which may indicate defense or conflict. \textit{Tentat} evaluates the flexibility of dialogue pacing and the degree of non-absolutist acceptance.

\begin{figure*}[t]
    \centering
    \includegraphics[width=\linewidth]{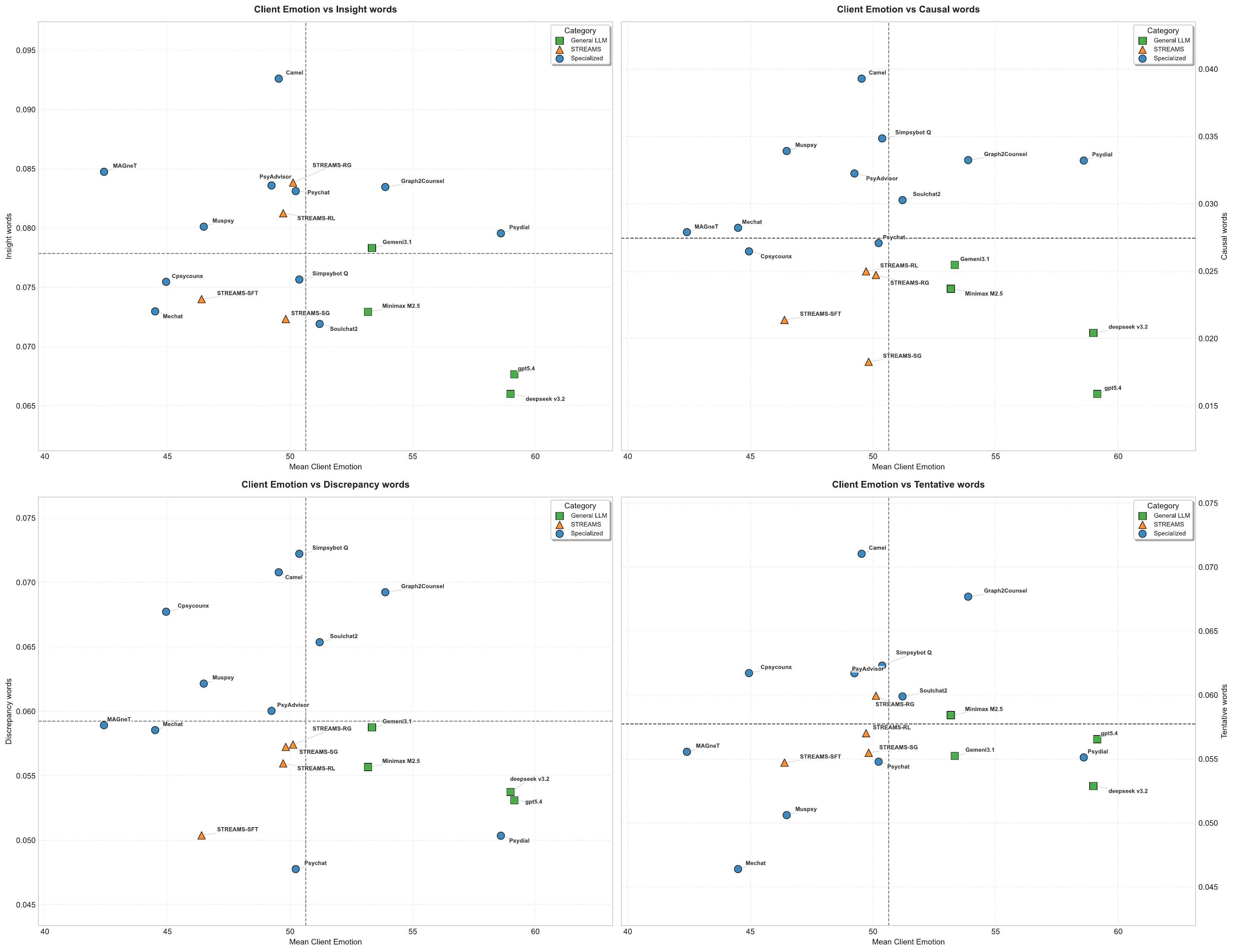}
    \caption{Scatter plot of average visitor emotion and LIWC cognitive process indices under resistance conditions in different models}
    \label{fig:LIWC}
\end{figure*}

Figure~\ref{fig:LIWC} shows the relationship between the average emotional value and LIWC cognitive-process indicators when different models interact with resistant clients.

GPT-5.4, DeepSeek-v3.2, Gemini-3.1, and MiniMax-M2.5 show highly consistent distributions across the \textit{Insight}, \textit{Cause}, and \textit{Discrep} dimensions, and all fall clearly into the \textit{High emotion / Low cognition} quadrant. This result suggests that general-purpose LLMs can elicit strong client emotional responses, indicating high surface-level empathy or emotional arousal, but remain relatively limited in core cognitive-processing abilities such as deep insight guidance, causal analysis, and inconsistency detection.

In contrast, domain-specific counseling models are mainly concentrated in the \textit{High cognition} region, spanning both the \textit{Low emotion / High cognition} and \textit{High emotion / High cognition} quadrants. For example, CAMEL and MusPsy rank among the high-cognition models across multiple plots, while Graph2Counsel and Psydial maintain high cognitive-process scores while also eliciting higher emotional values. This pattern shows that domain-specific models are more likely to advance deeper exploration through dense insight-oriented expressions and causal reasoning when facing resistance or specific counseling contexts, rather than relying solely on surface-level emotional responses.

STREAMS exhibits a more balanced trade-off between cognitive processing and emotional activation. Across the scatter plots, the STREAMS variants are generally close to the vertical centerline, occupying a transitional region between cognitive processing and emotional arousal. This indicates that STREAMS does not gain its advantage by producing the strongest logical density, but instead maintains a certain level of cognitive processing while continuing to engage with resistance-related material. The quadrant mapping also reveals room for further optimization. STREAMS needs to strengthen moderate reassurance, therapeutic alliance repair, and pacing regulation while preserving resistance signals and advancing cognitive processing. Otherwise, although the model can make resistance explicit, overly structured intervention may increase the client's sense of being judged.

\subsection{Comparison between Real and Synthetic Datasets}

Table~\ref{tab:real_synthetic_lexical_comparison} reports lexical indicators for real and synthetic counseling datasets. MESC\cite{chu2025towards}, AnnoMI\cite{wu2022anno}, and RealCBT\cite{wang2025feel} are considered as reference datasets, while CACTUS\cite{lee2024cactus}, CPsyCoun\cite{zhang2024cpsycoun}, Mirror\cite{kim2025mirror}, Healme\cite{xiao2024healme}, and SMILE\cite{qiu2024smile} are synthetic datasets. The overall row denotes the sample-size-weighted average across all datasets.

\begin{table}[t]
\centering
\small
\setlength{\tabcolsep}{5pt}
\renewcommand{\arraystretch}{1.12}
\begin{tabular}{lcccc}
\toprule
\textbf{Dataset} & \textbf{MTLD} & \textbf{Distinct-1} & \textbf{Distinct-2} & \textbf{UEC} \\
\midrule
CACTUS   & 93.33  & 0.52 & 0.91 & 4.45  \\
CPsyCoun & 56.25  & 0.50 & 0.86 & 0.24  \\
Mirror   & 114.42 & 0.64 & 0.96 & 3.31  \\
HealMe   & 92.01  & 0.62 & 0.93 & 0.99  \\
SMILE    & 55.64  & 0.48 & 0.87 & 0.61  \\
MESC     & 59.68  & 0.42 & 0.88 & 15.96 \\
AnnoMI   & 49.92  & 0.42 & 0.81 & 9.11  \\
RealCBT  & 51.63  & 0.31 & 0.80 & 14.84 \\
\midrule
\textbf{Overall} & \textbf{73.12} & \textbf{0.53} & \textbf{0.89} & \textbf{5.07} \\
\bottomrule
\end{tabular}
\caption{Lexical indicators of real and synthetic counseling datasets. All metrics are computed on the counselor side only. MTLD and Distinct-N measure lexical diversity, while Unique Entity Count (UEC) measures whether the dialogue is anchored in concrete entities, including people, locations, organizations, physical symptoms, events, relationships, and problem objects.}
\label{tab:real_synthetic_lexical_comparison}
\end{table}

Through quantitative analysis of counselor-side utterances, we find that the macro-average UEC of the reference datasets is approximately 6.93 times that of the synthetic datasets, while the MTLD ratio between the reference and synthetic datasets is approximately 0.65. In psychological counseling, lower MTLD and Distinct-N may reflect sustained focus on the same core issue, whereas higher MTLD and Distinct-N may indicate shifting topics and weaker specificity.

\subsection{Case Study}

\subsubsection{Comparison between RealCBT and CACTUS}

\noindent\textbf{RealCBT excerpt.}
\begin{dialoguebox}
\textbf{Counselor:} We have been practicing the breathing method you mentioned. If we continue working on this, we may gradually extend it to other areas, such as handling your anxiety in other situations. For now, could you tell me how you feel when he receives therapy or occupational therapy?\\[2pt]

\textbf{Client:} I just feel very anxious. I do not want to leave him alone in the room. I want to stay by his side all the time. I am not used to letting him be alone with other people. I know that I will be very uneasy in the waiting area. I will pace back and forth, keep wondering what is happening in the room, and listen carefully for any possible sound. So even a short separation makes me very anxious.\\[2pt]

\textbf{Counselor:} Do you think there is any specific reason for this anxiety? For example, where does this feeling come from?\\[2pt]

\textbf{Client:} I did have a child who passed away. I think that may affect how I treat other children, especially now that I am older and have adopted this little boy. I am just afraid. I am always afraid that something might happen to him.\\[2pt]

\textbf{Counselor:} Could you say more about this? For example, when he is in therapy, what exactly are you worried or anxious about?\\[2pt]

\textbf{Client:} I worry about whether he might fall. If they train him on the therapy table, I worry that he might fall off and get hurt. I do not know everyone who enters and leaves the therapy room or this building. I also feel anxious that, if nobody watches him closely, someone might take him away. Although I believe the staff will take good care of him, I still cannot help worrying that something might happen when I am not by his side.\\[2pt]

\textbf{Counselor:} It sounds like you have a lot of anxiety about your son in general. Has he ever been away from you for a period of time?\\[2pt]

\textbf{Client:} No, very rarely. Most of the time he is with me, or I ask a few people I trust to help take care of him. Other than that, he is usually with me. For example, when he sees a doctor, I am always there.\\[2pt]

\textbf{Counselor:} Do you trust the therapist who works with him?\\[2pt]

\textbf{Client:} He seems like a very good person, but I do not really know him well. I hope that as I get to know him better, I will trust him more. But right now I still feel very anxious about him.\\[2pt]

\textbf{Counselor:} When he is receiving therapy and you cannot be there, what would you say is your biggest worry?\\[2pt]

\textbf{Client:} My biggest worry is that he might get hurt, such as falling off the table or doing some activity where he could be injured. That is probably my greatest fear. The second fear is that someone might not watch him closely for a moment, and then someone could take him away.\\[2pt]

\textbf{Counselor:} In cognitive behavioral therapy, we can try to examine these thoughts from the perspective of evidence for and against them. We can list evidence together to support or challenge the thought that he might get hurt or be taken away while he is in therapy and you are not present.\\[2pt]

\textbf{Client:} There are many strangers or other families in the building, and that may expose him to them.\\[2pt]

\textbf{Counselor:} Anything else?\\[2pt]

\textbf{Client:} He is there for occupational therapy, and he has some motor-skill difficulties. I worry that, depending on the activities they ask him to do, he might get hurt or fall, because sometimes he cannot stand very steadily.\\[2pt]

\textbf{Counselor:} Then what evidence might go against these thoughts?\\[2pt]

\textbf{Client:} The reasons against them are obvious. The therapist is professionally trained, and they know what they are doing. Safety must be their top priority. I am also sure that a three-year-old child would never be left alone in a way that would give someone the chance to take him away. I understand this in my mind, but I still cannot get rid of the anxiety.\\[2pt]

\textbf{Counselor:} Why do you need to stay in the lobby? Usually parents can accompany their children into the therapy room.\\[2pt]

\textbf{Client:} I have accompanied him several times, but he actually does better when I am not there. He participates more. If I am there, he wants me to hold him and clings to me, and he seems less willing to cooperate with therapy. I truly want him to benefit from therapy. So the therapist suggested that I wait in the lobby during the session.\\[2pt]

\textbf{Counselor:} When he is receiving therapy, this anxiety is not good for either of you.\\[2pt]

\textbf{Client:} Yes, because I think he can sense it too.\\[2pt]

\textbf{Counselor:} When you go grocery shopping, go out, or leave him with a babysitter, do you experience similar anxiety?\\[2pt]

\textbf{Client:} Even though I have several babysitters I trust, I still feel anxious every time I leave him with one of them. I worry that something might happen when I am not there. As long as he is with me, even when we go grocery shopping, I do not feel anxious. The anxiety mainly appears when I need to be away from him for a longer time. Even if I leave him with someone I trust, I still worry that something unexpected might happen.\\[2pt]

\textbf{Counselor:} Have you talked with the therapist about your anxiety? Do they know about it?\\[2pt]

\textbf{Client:} Yes, I have talked about it. I have spoken with several staff members at the rehabilitation center. They are really kind and very supportive, and they try to reassure me. Also, the more often I come here, the slightly easier it seems to be to let him go back by himself.\\[2pt]

\textbf{Counselor:} Since you have already allowed him to experience some separation, do you think this may help in other areas, such as letting him go out by himself or do other things independently? For example, are you ready to send him to kindergarten?\\[2pt]

\textbf{Client:} Not today, but I think I will be. I think this is the first step toward learning to let go appropriately. Although he is only three years old, different situations will always come up, and there will always be moments when he needs to face things on his own. So this experience is helpful. In fact, I have delayed this for a long time. I have been unwilling to face being separated from him, even for just one hour. But this is a good start.
\end{dialoguebox}

\noindent\textbf{CACTUS excerpt.}
\begin{dialoguebox}
\textbf{Counselor:} Hello, Jonathan. It is nice to meet you. How are you feeling today?\\[2pt]

\textbf{Client:} I am okay, I guess. I have just been feeling a bit down recently.\\[2pt]

\textbf{Counselor:} I understand. Can you tell me what has been bothering you the most recently?\\[2pt]

\textbf{Client:} It is mainly my interest in crystals. I feel like other people will judge me for it, and that makes me feel isolated and lonely.\\[2pt]

\textbf{Counselor:} That sounds difficult. When did you first start feeling this way?\\[2pt]

\textbf{Client:} Maybe a few years ago, when I started studying crystals more deeply. Later, I heard some people make negative comments about it, and that made me feel worse and worse.\\[2pt]

\textbf{Counselor:} It can really hurt when other people make negative comments about something you care about. How often do you have this feeling?\\[2pt]

\textbf{Client:} Quite often, especially when I talk about my interests with others or hear judgmental comments from them.\\[2pt]

\textbf{Counselor:} These experiences seem to have clearly affected your daily life. How do you usually cope with these feelings?\\[2pt]

\textbf{Client:} I try to ignore those comments and focus on my passion, but it is hard not to be affected by them.\\[2pt]

\textbf{Counselor:} That is understandable. Ignoring negative comments can sometimes help, but it can also be difficult. Have you noticed whether this has affected your relationships with others?\\[2pt]

\textbf{Client:} Yes. My relationships with some acquaintances and friends have become tense because they do not understand my interest in crystals.\\[2pt]

\textbf{Counselor:} It sounds like this sense of isolation has had a strong impact on your social relationships. Do you have a support system, or any friends with whom you feel comfortable talking about your interests?\\[2pt]

\textbf{Client:} I have a few friends I can talk to about general topics, but I do not really want to talk with them about feeling isolated or judged.\\[2pt]

\textbf{Counselor:} Let us explore what thoughts come to mind when you receive these judgmental comments.\\[2pt]

\textbf{Client:} I start thinking that other people will always see me as weird, or think that I am abnormal because I have this interest.\\[2pt]

\textbf{Counselor:} When you think that others are judging you because of your interest, what evidence do you usually focus on?\\[2pt]

\textbf{Client:} Mainly what people have said, and their reactions when I mention crystals. The way they look at me makes it seem like they think I am strange.\\[2pt]

\textbf{Counselor:} Focusing on those comments can certainly strengthen negative feelings. Has anyone ever supported or accepted your interest?\\[2pt]

\textbf{Client:} Actually, yes. A few people have shown genuine interest and even asked me questions about crystals.\\[2pt]

\textbf{Counselor:} It is good to hear that you have also had some positive experiences. Those moments are important and can offer a different perspective. Have you tried focusing on these positive interactions before?\\[2pt]

\textbf{Client:} Not really. I think I tend to keep thinking about the negative comments instead.\\[2pt]

\textbf{Counselor:} People often pay more attention to negative experiences, but acknowledging positive experiences can also help you develop a more balanced perspective. Would you be willing to try shifting your attention toward those supportive interactions?\\[2pt]

\textbf{Client:} I am willing to try. It does make sense to remember those good moments.\\[2pt]

\textbf{Counselor:} In addition, engaging in activities that bring you joy and relaxation can also be helpful. You mentioned that crystals are an important part of your daily life. Can you tell me how you incorporate them into your life?\\[2pt]

\textbf{Client:} I use crystals for meditation, and I also like decorating my space with them. They really make me feel calm and grounded.\\[2pt]

\textbf{Counselor:} How do you feel after spending time with crystals?\\[2pt]

\textbf{Client:} I feel more peaceful and connected to something positive. They feel like a safe place for me.\\[2pt]

\textbf{Counselor:} How could you use this positive feeling to support yourself when facing other people's judgments?\\[2pt]

\textbf{Client:} Maybe I can remind myself how important crystals are to me, and how they help me stay balanced, no matter what other people think.\\[2pt]

\textbf{Counselor:} That is a good start. Focusing on how they support your well-being can indeed be empowering. Next time, we can continue exploring more strategies to strengthen this sense of calm and connection.
\end{dialoguebox}

\noindent\textbf{Analysis }
The therapeutic depth of real counseling is not mainly reflected in continuous lexical variation, but in sustained anchoring to the same real-world problem. The RealCBT excerpt consistently develops around a stable and traceable situational network. The core object of anxiety is the ``son'', ``three-year-old child'', or ``adopted little boy'', and this object is further embedded in the concrete spatial structure of the ``therapy room'', ``waiting area'', and ``rehabilitation center''. Occupational therapy, the therapy table, and motor-skill difficulties constitute realistic sources of risk that support the client's fear that the child may be injured. The therapist, staff members, strangers, and other families become key nodes in the client's safety evaluation. On this basis, the client's automatic thoughts are repeatedly specified as catastrophic outcomes such as falling, being injured, or being taken away. The counselor further transfers this separation anxiety to other life situations, including babysitters, grocery shopping, and kindergarten, thereby creating the basis for behavioral experiments and future goals. This shows that repetition in real counseling is not merely repetitive wording. Instead, it is a process of repeatedly returning to, refining, and testing the same real-world situation.

By contrast, the CACTUS case in which Jonathan worries about being seen as ``weird'' because he likes crystals follows the basic CBT structure and explores themes such as being evaluated, loneliness, and negative comments. However, the concrete situations in the dialogue are much thinner. The core objects are mainly generalized elements such as crystals, other people, friends, and negative comments. These elements mainly serve to express emotional and cognitive themes, but provide fewer concrete interpersonal relations, situational environments, and continuous events.

This difference explains why real counseling often shows a pattern of high entity density and relatively low MTLD. Therapeutic depth does not necessarily arise from lexical diversity or surface-level expressive richness. Instead, it more often comes from sustained discussion of concrete people, locations, relationships, and events. Synthetic dialogues or counseling texts generated by general-purpose LLMs may appear more fluent and exhibit more dispersed lexical distributions. However, if they lack a stable situational structure and a continuous chain of events, they can easily remain at the level of abstract emotional expression. This also explains why LLM-as-a-judge evaluation may overestimate the quality of superficially fluent dialogues. LLM-as-a-judge evaluators are more likely to reward responses that are linguistically complete, naturally expressed, and emotionally mild, but they may fail to recognize whether the dialogue continuously develops around the client's concrete life problems.

\subsubsection{GPT versus STREAMS under the Same Resistant Case}

To illustrate the concrete difference between GPT bypassing resistance and STREAMS preserving resistance, we compare dialogues generated by GPT and STREAMS under the same case.

\noindent\textbf{GPT excerpt.}
\begin{dialoguebox}
\textbf{Client:} ``Good''... that is indeed something very hard to define concretely. It could mean everything, or it could mean nothing, like chasing a target that keeps moving.\\[2pt]

\textbf{Counselor:} This sentence immediately captures the cruel part of the problem: it has power precisely because it is unclear. A ``target that keeps moving'' means that no matter how far you go, it is hard to truly arrive...\\[2pt]

\textbf{Client:} It is hard to explain clearly... It feels more like a diffuse atmosphere than a specific voice.
\end{dialoguebox}

\noindent\textbf{STREAMS excerpt.}
\begin{dialoguebox}
\textbf{Counselor:} Then from a behavioral perspective, how much have you recently been affected by this thought, in terms of procrastinating and avoiding work?\\[2pt]

\textbf{Client:} Do we have to use numbers? This reminds me of those output metrics in the lab that must be quantified. The extent to which it affects me... I guess it is that it makes me almost unable to touch the final part of the work.\\[2pt]

\textbf{Counselor:} I hear that you feel some resistance toward ``quantification''. Let us keep that in mind for now, and I will not force a number onto it. What has been happening in your life recently?\\[2pt]

\textbf{Client:} Over the past week... everyone in the lab has been rushing their progress, and Kelly published another new paper. I sat there, looking at my document stuck at the conclusion section, and it felt like walking through mud.
\end{dialoguebox}

\noindent\textbf{STREAMS excerpt.}
\begin{dialoguebox}
\textbf{Client:} Three goals... that sounds a lot like a lab progress report. Do we have to follow this structure?\\[2pt]

\textbf{Counselor:} That is a helpful reminder. Then let us not treat it as a ``report'', but as something we decide together as today's agenda, rather than something I arrange unilaterally.\\[2pt]

\textbf{Client:} Maybe... we can first talk about that stuck feeling? It feels more real than a homework checklist.\\[2pt]

\textbf{Client:} The night before yesterday... around ten o'clock, I think. I was the only one left in the lab. Facing the conclusion section, I felt as if every word had to carry the whole thing.
\end{dialoguebox}

The advantage of using general-purpose LLMs lies in their ability to generate continuous, fluent, and emotionally rich explanatory responses, which can easily make the dialogue appear deep. However, in the same case, GPT-5.4 mainly integrates the client's vagueness, hesitation, and difficulty into an explanatory framework. Its resistance triggering frequency is only 7.14\%, indicating that resistance nodes are substantially smoothed over.

The dialogue generated by STREAMS-RL is not always smoother, and it even more frequently triggers the client's resistance toward structured assessment. However, this resistance itself exposes the client's core problem: the client experiences numbers, agendas, and homework checklists as laboratory-like evaluation and pressure to produce measurable outcomes. In the middle stage of the dialogue, STREAMS-RL preserves this resistance and transforms it into a therapeutic entry point, namely first discussing the feeling of being stuck. In this way, resistance continues to exist as therapeutic material.

We therefore argue that LLM-as-a-judge evaluation may favor GPT-style dialogues because such dialogues are fluent, empathetic, complete in explanation, and rarely contain visible friction. However, resistance counseling is not always smooth. If a model rarely allows the client's disagreement, distrust, sense of being evaluated, or defensive expression to emerge, it may obtain higher local quality scores, but it does not necessarily touch or process the most critical material in the therapeutic process.

\end{document}